\documentclass[10pt,journal,compsoc]{IEEEtran}
\usepackage{lineno,hyperref}
\modulolinenumbers[5]
\usepackage{times}
\usepackage{epsfig}
\usepackage{graphicx}
\usepackage{amssymb}
\usepackage{amsmath}
\usepackage{amsmath }
\usepackage{multirow}
\usepackage{xcolor}
\usepackage{hhline}
\usepackage{algorithm}
\usepackage{algpseudocode}
\usepackage{caption}
\usepackage{array}
\usepackage{placeins}
\usepackage{booktabs}
\usepackage{empheq}
\usepackage{graphicx}
\usepackage{amsthm}
\usepackage{amsmath}
\usepackage{multirow}
\usepackage{multicol}
\usepackage{enumitem}
\usepackage{svg}
\usepackage{color, colortbl}
\usepackage[flushleft]{threeparttable}
\DeclareMathOperator*{\argmax}{arg\,max}
\DeclareMathOperator*{\argmin}{arg\,min}

\newtheorem{proposition}{Proposition}

\definecolor{Gray}{gray}{0.9}
\newcolumntype{g}{>{\columncolor{Gray}}c}
\definecolor{m_purple}{HTML}{6F2DBD}
\definecolor{m_green}{HTML}{2C8915}
\definecolor{r_impro}{HTML}{6AAA96}
\definecolor{r_impro_n}{HTML}{E67F83}

\begin{document}
\title{Iterative Knowledge Exchange Between Deep Learning and Space-Time Spectral Clustering for Unsupervised Segmentation in Videos}

%
%

\author{Emanuela~Haller,~
        Adina~Magda~Florea,~\IEEEmembership{Senior Member,~IEEE,}
        and~Marius~Leordeanu
\IEEEcompsocitemizethanks{\IEEEcompsocthanksitem E. Haller is with University Politehnica of Bucharest and Bitdefender.\protect\\
E-mail: haller.emanuela@gmail.com
\IEEEcompsocthanksitem A.M. Florea is with University Politehnica of Bucharest.\protect\\
E-mail: adina.florea@gmail.com
\IEEEcompsocthanksitem M. Leordeanu is with University Politehnica of Bucharest, Institute of Mathematics of the Romanian Academy and Bitdefender.\protect\\
E-mail: leordeanu@gmail.com
}
\thanks{(Corresponding author: M. Leordeanu.)}}

\IEEEtitleabstractindextext{%

\begin{abstract}
We propose a dual system for unsupervised object segmentation in video, which brings together two modules with complementary properties: a space-time graph that discovers objects in videos and a deep network that learns powerful object features. The system uses an iterative knowledge exchange policy. A novel spectral space-time  clustering process on the graph produces unsupervised segmentation masks passed to the network as pseudo-labels. The net learns to segment in single frames what the graph discovers in video and passes back to the graph strong image-level features that improve its node-level features in the next iteration. Knowledge is exchanged for several cycles until convergence. The graph has one node per each video pixel, but the object discovery is fast. It uses a novel power iteration algorithm computing the main space-time cluster as the principal eigenvector of a special Feature-Motion matrix without actually computing the matrix. The thorough experimental analysis validates our theoretical claims and proves the effectiveness of the cyclical knowledge exchange. We also perform experiments on the supervised scenario, incorporating features pretrained with human supervision. We achieve state-of-the-art level on unsupervised and supervised scenarios on four challenging datasets: DAVIS, SegTrack, YouTube-Objects, and DAVSOD. 
\end{abstract}

\begin{IEEEkeywords}
Unsupervised video object segmentation,
spectral space-time clustering, principal eigenvector, power iteration, space-time graph, Feature-Motion matrix, object discovery, unsupervised learning, deep learning.
\end{IEEEkeywords}}

\maketitle

\IEEEdisplaynontitleabstractindextext

%

\IEEEraisesectionheading{\section{Introduction}\label{sec:introduction}}

\IEEEPARstart{D}{iscovering} objects, without human supervision, as they move and change appearance over space and time is one of the most challenging and still unsolved problems in computer vision. One of the main questions we ask is how we could best exploit the correlation between objects' motion and appearance to model mathematically the process of object discovery in the absence of human supervision. By solving this task, we could learn more efficiently from large quantities of data available in the space-time domain, with minimal human intervention.

The task of visual grouping, which comes naturally to human beings, proves to be very demanding for machines. We learn by observing the surrounding world, and, while nobody explicitly teaches us how, we are so good at it. An interesting aspect is that we strongly agree on the main foreground object shown in a given video shot. While visual grouping has been studied since the early Gestalt School of psychology \cite{koffka2013principles}, we have not yet found the underlying computational model that could best describe such unsupervised discovery. We hope that the approach and mathematical formulation proposed in this article could shed new light on solving this very interesting task.
In the context of unsupervised segmentation in video,
we bring face-to-face two important computational learning worlds: that of \textbf{deep learning}, with strong supervised learning abilities, and that of \textbf{iterative graph algorithms}, with proved unsupervised clustering strengths. On the one hand, feed-forward neural networks are powerful if trained on large quantities of high-quality labeled data. On the other hand, classical graphical models can transcend their initialization point through iterative message-passing schemes when given the right functions and features on their nodes and edges. In the proposed iterative knowledge-exchange paradigm, we put graphs, and deep nets face to face. The graph becomes the teacher, providing pseudo-ground truth supervisory signal to the deep network. The deep network then learns powerful features that are passed back to the graph to improve the graph model. The two modules complement each other and exchange knowledge for several cycles until convergence. Please note that, as also discussed in the related work section, our dual graph-network model is very different from the well-known graph neural networks literature and has distinctive contributions in unsupervised learning. For more details, code, and demo results, visit the project website\footnote{\url{https://shorturl.at/oBOS8}}.

We introduce a space-time graph structure that can be exploited to automatically segment the primary object of a video sequence in an unsupervised setup.
While general 3D convolution-based approaches (\cite{tran2015learning, carreira2017quo}) treat the time dimension as being equivalent to the spatial ones, we propose a different way of coupling motion and appearance. Objects in the real world form clusters in their space-time neighborhoods, and points belonging to the same object stay connected in both space and time, having similar appearance and motion patterns that are also distinctive from the rest of the scene. These real-world properties immediately translate in the video domain since each frame is a 2D projection of the world. We observe that video pixels of the same object are highly likely: \textbf{(1)}~to be connected through long-range motion chains, created by successive optical flow vectors connecting pixels from one frame to the next (Sec.~\ref{sec:graph_module_spacetime_graph}); \textbf{(2)}~to be distinctive regarding both appearance and motion patterns in space and time. Thus, pixels that look alike and move together in a video sequence are likely to belong to the same object, forming a strong and distinctive space-time cluster. Based on this intuition, we construct the graph using a novel Feature-Motion adjacency matrix. Our key claim is that the main cluster of the graph, within a certain space-time neighborhood (as defined by the video shot), naturally corresponds to the most salient object in the video, a fact that is supported by our extensive experiments. Also, in order to compute this primary cluster fast, we introduce a novel algorithm for spectral clustering, which finds the cluster on the principal eigenvector of the Feature-Motion matrix, without explicitly going through the expensive process of building the matrix (Sec.~\ref{sec:graph_module_optimization_alg}). 

In its basic formulation, the graph does not use pretrained features, which could limit its representation power at the first cycle. However, the alternation between the graph and the network modules helps overcome the limitation, as they exchange knowledge alternatively and play teacher and student roles until convergence. While the graph exploits the space-time consistency, the network builds powerful object features by
learning (using the graph's output as pseudo-ground truth) to segment the object in single frames. The dual system benefits from the complementarity between graph and neural net in a novel and efficient way, such that each module takes advantage of its own strengths (graph - the ability to discover clusters unsupervised; deep net - the ability to learn powerful deep features) to address the other's weaknesses (graph - the need for powerful node-level features; deep net - the need for supervision). Together, they form a strong team that can function as a single self-supervised entity.

    \begin{figure*}[t]
    \includegraphics[width=1\textwidth]{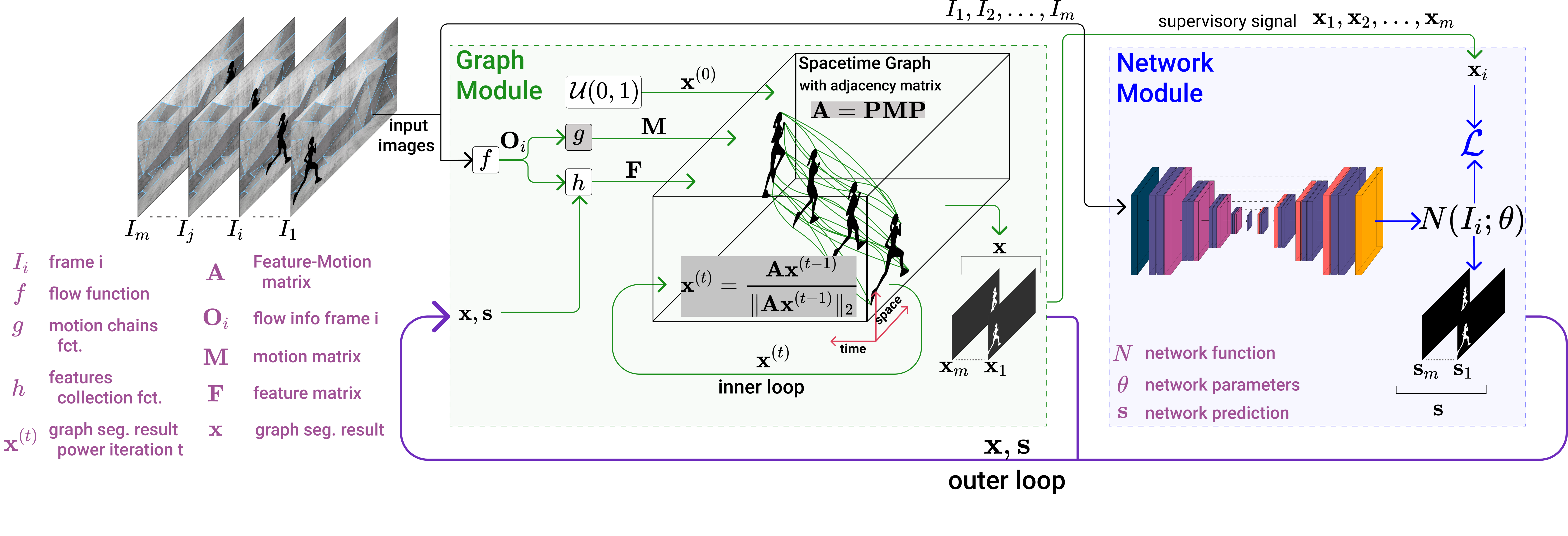}
    \centering
    \caption{Architecture of our Iterative Knowledge Exchange (\textbf{IKE}) system. The \textcolor{m_green}{Graph Module} (left) and the \textcolor{blue}{Network Module} (right), exchange information over several cycles (outer loops) until convergence (Alg.~\ref{alg:cs-gdm}). The \textcolor{m_green}{Graph Module} discovers the main object in the video as the strongest cluster of the \textcolor{m_green}{space-time graph} with one-to-one correspondences between graph nodes and video pixels. The segmentation solution is the principal eigenvector of a novel Feature-Motion matrix ($\mathbf{A}$) computed efficiently with a specially designed power iteration algorithm (Alg. \ref{alg:practical_iterations}: inner loop of Graph Module). Then, the \textcolor{blue}{Network Module} is trained from scratch, using the segmentation result of the graph as pseudo-ground truth. The computed deep representations ($\mathbf{s}$), along with the graph segmentation results ($\mathbf{x}$) are used to complete node-level features and reinitialize the space-time graph model for a new cycle (outer loop). We emphasize that the gray-colored functions ($g$, power iteration equation, and computation of adjacency matrix $\mathbf{A}$) are only conceptual entities, as the actual algorithm avoids explicitly computing $\mathbf{M}$ and $\mathbf{A}$, drastically reducing the computational burden and memory usage (Sec.~\ref{sec:graph_module_optimization_alg}). In Sec.~\ref{sec:graph_module_th_analysis}, we provide theoretical guarantees that our power iteration equivalent algorithm indeed converges to the main eigenvector of $\mathbf{A}$, which is the optimal solution of the relaxed problem (Eq. \ref{eq_problem1}). In extensive tests, the proposed \textbf{IKE} system significantly boosts its performance over several cycles (Fig.~\ref{fig:CS_GDM_results}, Table~\ref{tab:full_system_avgs}). [Best viewed in color]}
    \label{fig:cs_gdm}
    \end{figure*}
    
    The \textbf{main contributions} of our work are: \textbf{(1)} We introduce a dual \textbf{Iterative Knowledge Exchange (IKE)} system  (Sec.~\ref{sec:approach}, Fig.~\ref{fig:cs_gdm}), which puts together a space-time graph and a deep neural network, to learn, as a single self-supervised entity. The two modules alternatively share their learned knowledge until reaching an equilibrium at convergence. The extensive tests in both unsupervised and supervised scenarios with state-of-the-art results in almost all cases on four challenging datasets also show the value of our approach;
    \textbf{(2)} The \textbf{Space-time Graph Module} (Sec.~\ref{sec:graph_module}), with one node per video pixel, based on the Feature-Motion Matrix is completely novel, along with the fast algorithm for object discovery in video formulated as space-time spectral clustering. We prove that the algorithm, which computes the principal eigenvector without explicitly building the matrix, is equivalent to power iteration applied to the Feature-Motion matrix.

\begin{algorithm*}
        \caption{main algorithmic steps of our \textbf{IKE} system (Fig.~\ref{fig:cs_gdm})\\ 
        \textit{Note:} in our implementation (Sec \ref{sec:graph_module_optimization_alg}), steps 3, 6 and 9 avoid the actual computation of $\mathbf{M}\text{ and }\mathbf{A}$ (their usage is implicit).}
        \label{alg:cs-gdm}
        \begin{algorithmic}[1]
            \Repeat 
            \Statex
            \Statex{\textbf{\textcolor{m_green}{Graph Module}}} \Comment (Sec.~\ref{sec:graph_module})
            
            \State for each frame $i$: $\mathbf{O}_i = f(I_{i-1}, I_i, I_{i+1})$
            \Comment compute forward and backward optical flow

            \State $\mathbf{M} = g(\mathbf{O}_1, ...\mathbf{O}_m)$
            \Comment define space-time graph motion matrix (Sec.~\ref{sec:graph_module_spacetime_graph}, Fig.~\ref{fig:spacetime_graph})
            
            \State for each frame $i$: $\mathbf{\Gamma}_i \leftarrow (\mathbf{O}_i, \mathbf{x}_i, \mathbf{s}_i)$
            \Comment get pixel-level frame feature maps, for current cycle
            
            \State $\mathbf{F} \leftarrow h(\mathbf{\Gamma}_1, ..., \mathbf{\Gamma}_m)$ 
            \Comment collect node-level features and define feature matrix  (Sec.~\ref{sec:graph_module_spacetime_graph}, Fig.~\ref{fig:chain_features})
            
            \State $\mathbf{A} \leftarrow \mathbf{P}\mathbf{M}\mathbf{P} \text{ with } \mathbf{P}=\mathbf{F}(\mathbf{F}^T\mathbf{F})^{-1}\mathbf{F}^T$
            \Comment define Feature-Motion matrix (Sec.~\ref{sec:graph_module_problem_formulation})
            
            \State for each pixel $i$: $x_i \stackrel{iid}{\sim} \mathcal{U}(0,1) \Rightarrow \mathbf{x}^{(0)}$
            \Comment initialize node labels  (Sec.~\ref{sec:exp_analysis_graph_module_conv})
            
            \Repeat
            \State{$\mathbf{x}^{(t)} = \frac{\mathbf{A}\mathbf{x}^{( t-1)}}{\|\mathbf{A}\mathbf{x}^{(t-1)}\|_2}$}            
            \Until{convergence to optimal solution $\mathbf{x}$} \Comment power iteration algorithm
            (Sec.~\ref{sec:graph_module_optimization_alg})
            
            \Statex
            \Statex{\textbf{\textcolor{blue}{Network Module}}} \Comment (Sec.~\ref{sec:network_module})
            
            \State $D=\{(\mathbf{I}_i, \mathbf{x}_i) | i\in\{1, ..., m\}\}$
            \Comment build training set, using the pseudo-ground truth from the Graph Module\\
            \Comment video-level solution $\mathbf{x}=(\mathbf{x}_1,...,\mathbf{x}_m)$, where $\mathbf{x}_i$ is the segmentation of frame i
            
            \State $\theta^*=\argmin_\theta \frac{1}{m}\sum_{i=1}^m{\mathcal{L}(N(\mathbf{I}_i;\theta), \mathbf{x}_i)} $
            \Comment train the deep segmentation model, from scratch
            
            \State $\mathbf{s}_i \leftarrow N(\mathbf{I}_i,\theta^*)$ 
            \Comment compute network segmentation predictions $\mathbf{s}=(\mathbf{s}_1, ..., \mathbf{s}_m)$\\
            
            \Until{reaching convergence of Graph and Network Modules}
       
        \end{algorithmic}
    \end{algorithm*}
    
\section{Scientific context}\label{sec:scientific_context}

   Video object segmentation is gaining rapid development in computer vision. Most solutions are fundamentally supervised, as they rely on heavily pretrained models with human-labeled annotations~\cite{zhou2020matnet, luiten2018premvos, maninis2017video, voigtlaender2017online, bao2018cnn, wug2018fast, cheng2018fast, caelles2017one, perazzi2017learning, chen2018blazingly, song2018pyramid, tokmakov2017learning, jain2017fusionseg}. 
    Although manual annotation is extremely costly, there are very few genuine unsupervised methods~\cite{lao2018extending, papazoglou2013fast, keuper2015motion, faktor2014video, haller2017unsupervised}. While the task has various formulations, here we focus on \textbf{zero-shot video object segmentation}, whose purpose is to discover and segment the main object in a video sequence, when only the raw video frames are provided as input, without any humanly supervised information provided. We mention that the main object could sometimes be composed of multiply strongly related ones, behaving as a whole (e.g., a man riding a bicycle).
    
    \noindent \textbf{Relation to Classical and Deep Learning Approaches:} More traditional solutions rely on different heuristic assumptions and auxiliary tasks, such as the computation of motion boundaries ~\cite{papazoglou2013fast} and "objectness" measure \cite{faktor2014video, li2017primary}. Recent approaches propose a supervised combination between motion and appearance~\cite{tokmakov2017learning_, li2018unsupervised}. Some propose a two-stream supervised model \cite{jain2017fusionseg, tokmakov2017learning}. In the recent MATNet model~\cite{zhou2020matnet} motion and appearance are deeply interleaved, allowing a richer representation power. Another interesting approach~\cite{lu2019see} exploits a specific co-attention mechanism to capture global correlations by analyzing multiple frame pairs in a Siamese network scheme. Different from the works mentioned above, ours bridges the gap between more classical iterative graph algorithms and deep learning in order to achieve self-supervision by exploiting the advantages of both.
    
    \noindent \textbf{Relation to Graph Neural Networks:} We bring together graphs and deep nets, an association that 
    could immediately remind us of graph neural networks ~\cite{scarselli2008graph,nicolicioiu2019recurrent}. Indeed, our system has some conceptual elements that are related graph neural network: the iterative process of our space-time graph can be associated to the neighbourhood aggregation process in GNNs~\cite{zhou2018graph} and our features collection (Fig.~\ref{fig:chain_features}) can be related to the features representation concatenation of GNNs~\cite{hamilton2017inductive}. Yet, our system differs significantly in terms of both architecture and learning abilities. Regarding architecture, in the case of GNNs, tiny networks (MLPs) sit inside the graph, on its nodes and edges, and model the message passing between the nodes. In our case, the graph and the large deep network face each other and function in tandem, exploiting each other's complementary learning properties. Moreover, our space-time graph is defined at the dense pixel level, which is again very different from the GNNs models that have a sparse grid-level structure, with only a few nodes per image, each with a dedicated large region (\cite{nicolicioiu2019recurrent}). Besides the clear structural differences between GNNs and the model proposed here, we believe that the most relevant difference is in the fact that our dual graph-deep network system is designed with unsupervised learning in mind and it fully exploits the unsupervised clustering capacity of the space-time graph by focusing on a powerful and efficient spectral approach. We should note that there is no work in GNN literature to the best of our knowledge that does unsupervised learning in vision. However, despite the evident differences, we find the idea of exploring the relationship between the two approaches to be an interesting potential future direction.

    \noindent \textbf{Relation to Methods using Optical Flow:} Several notable works use optical flow for video segmentation. One of the best known models~\cite{brox2010object} exploits motion clustering to simultaneously predict flow and object segmentation. They use an energy function that combines optical flow constraints with an appearance term. Unlike ours, that approach has no immediate mechanism for identifying the primary object in the video. A similar approach is proposed in~\cite{tsai2016video}. Another related work~\cite{zhuo2018unsupervised} builds salient motion masks, further combined with objectness to generate the final segmentation. Others~\cite{li2017primary,wang2015saliency} combine saliency and optical flow by computing an average of the saliency masks based on direct optical flow connections. Motion video segmentation is tackled in~\cite{keuper2015motion}, where different scene components are separated as different motion clusters. The task is formulated as a minimum cost multicut problem over a graph that considers nodes as corresponding to different motion trajectories. Two key elements differentiate our approach from all: \textbf(1) we propose a compact mathematical model, which couples motion and appearance, to define the primary object in the video as the main natural spectral cluster in our Feature-Motion matrix; \textbf(2) our space-time cluster is dense, at the pixel level and thus is able to use all information in the video without losing details by making hard grouping decisions early (e.g., computing superpixels). The fast power iteration method we introduce makes the formulation at the dense pixel level possible.
    
    \noindent \textbf{Relation to works using spatio-temporal consistency:} Our space-time graph exploits the temporal consistencies inherent in the video sequence. This temporal continuity plays a significant role in object recognition development. \cite{wood2016smoothness} presented a study performed on newborn chicks and showed that those raised with virtual-objects that move non-smoothly developed poorer object recognition abilities than those raised with virtual objects that moved smoothly. Recent works~\cite{wang2019learning, jabri2020space} exploit spatio-temporal graph structures inside self-supervised systems that learn object representations. While they effectively learn an embedding space by performing forward-backward passes through the video sequence, their formulation is very different from our approach of unsupervised space-time spectral clustering for segmentation.

    \begin{figure*}[ht]
    \includegraphics[width=1\textwidth]{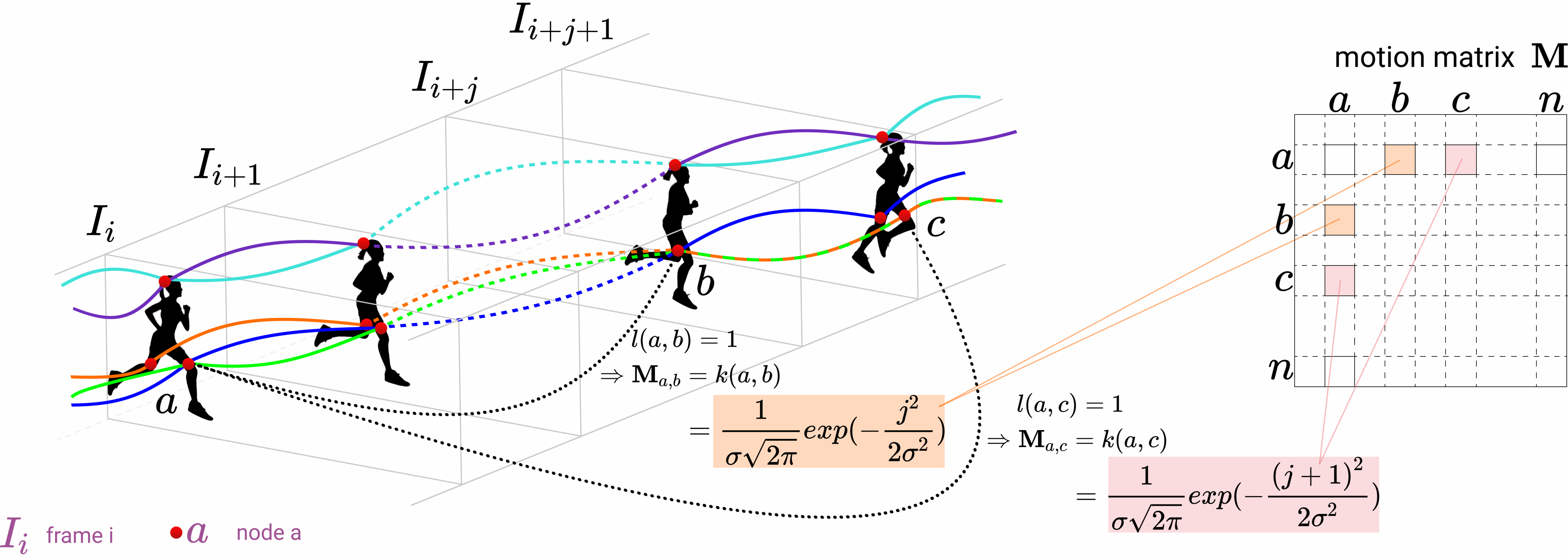}
    \centering
    \caption{Visual representation of the space-time graph structure. We illustrate the process of creating long-range edges that define our graph. \textbf{Colored curves} indicate motion chains, formed by following the optical flow vectors, forward and backward in time, sequentially from one frame to the other. \textbf{Black dotted curves} correspond to graph edges, defined between nodes connected through at least one motion chain. Consequently, two nodes are connected if and only if at least one chain of optical flow vectors links them through consecutive intermediate frames. Precisely two flows are going out from a given node (one in each direction), but there could be any number of incoming chains from any direction reaching a node (including none). The motion chain weight associated with each existing edge is a Gaussian kernel of the inter-frame (temporal) distance between its vertices. The set of edges, along with their weights, construct the graph motion matrix $\mathbf{M}$. Based on $\mathbf{M}$ and the feature matrix $\mathbf{F}$, we define the final graph adjacency matrix, termed the Feature-Motion matrix $\mathbf{A=PMP}$ (Sec.~\ref{sec:graph_module_problem_formulation}). [Best viewed in color]}
    \label{fig:spacetime_graph}
    \end{figure*}
    
\section{Approach}\label{sec:approach}
    
    We propose a dual Iterative Knowledge Exchange (\textbf{IKE}) model coupling space-time spectral clustering with deep object segmentation, able to learn without any human annotation. The graph module exploits the spatio-temporal consistencies inherent in a video sequence but has no access to deep features. The deep segmentation module exploits high-level image representations but requires a supervisory signal. The two components are complementary and together form an efficient self-supervised system  able to discover and learn to segment the primary object of a video sequence. The space-time graph is the first to play the teacher role by discovering objects unsupervised. Next, the deep model is trained from scratch for each video sequence, using the graph's output as pseudo-ground truth. At the second cycle, the graph module incorporates knowledge from the segmentation model, adding the powerful deep learned features that correspond to each of its nodes. The process in which the graph and the deep net exchange knowledge in this manner repeats over several cycles until convergence is reached. In Fig.~\ref{fig:cs_gdm}, we present the architecture of the proposed system, highlighting its two main modules: Graph Module and Network Module. 

    The Graph Module (Sec.~\ref{sec:graph_module}) is a central element  of our system and a key contribution. It defines unsupervised object segmentation as a clustering problem in the space-time graph associated with the video sequence. Unlike other similar approaches, we propose a dense representation, with one-to-one correspondences between video pixels and graph nodes. The main object is the strongest cluster in terms of both motion patterns and consistencies of the node-level features. In this formulation, the segmentation solution is the leading eigenvector of the Feature-Motion matrix, which is the space-time graph's adjacency matrix. We propose an iterative algorithm that is guaranteed to converge to the optimal solution, and we prove this both theoretically and experimentally. In Sec.~\ref{sec:graph_module}, we will formally introduce our method along with a theoretical analysis of the proposed approach.
    
    The Network Module (Sec.~\ref{sec:network_module}) comes as a complement of the Graph Module, adding deep features into the clustering algorithm. The network comes with a great representation power and attempts to predict the space-time clustering process's output having only single frames as input. This constraint forces the network to learn powerful high-level image features, which can be further exploited in the next cycle to enrich the space-time graph's node-level features. Thus, the Network Module's learning process is guided by our space-time graph's segmentation result and requires no human supervised pre-training process. In Sec.~\ref{sec:network_module}, we will introduce in detail the proposed module. 
    
    In Algorithm~\ref{alg:cs-gdm}, we outline the iterative knowledge exchange system's conceptual steps. Each step is presented by its equivalent mathematical formulation, and further details are introduced in Sec.~\ref{sec:graph_module} and Sec.~\ref{sec:network_module}. 
    
\section{Graph Module}\label{sec:graph_module}

    Given a sequence of $m$ consecutive video frames, the Graph Module discovers the primary object as the strongest natural cluster in the space-time graph and extracts a set of $m$ soft-segmentation masks, one per frame, corresponding to that main object. In Sec.~\ref{sec:graph_module_spacetime_graph}, we introduce the space-time graph, defined over the whole spatio-temporal volume of the video sequence, with one-to-one correspondences between graph nodes and video pixels and edges defined by motion chains. In Sec.~\ref{sec:graph_module_problem_formulation}, we formulate segmentation as a clustering problem in the space-time graph. Thus, the segmentation mask associated with a given frame is a slice of the identified space-time cluster in the entire video. Sec.~\ref{sec:graph_module_optimization_alg} presents the optimization algorithm, which efficiently computes the cluster as the leading eigenvector of the graph's adjacency Feature-Motion matrix. Then, in Sec.~\ref{sec:graph_module_th_analysis},  we provide the theoretical analysis with a formal proof that our algorithm indeed converges to the leading eigenvector of the Feature-Motion matrix, even though the matrix itself is never built explicitly, which would be very expensive.
    
    \subsection{Space-time
    Graph}\label{sec:graph_module_spacetime_graph}
    
    We define the \textbf{space-time graph} $G=(V,E)$, with a node $a\in V$ associated to each pixel of the video ($|V|=n$, where $n=mhw$, with $m$ the number of frames and $(h,w)$ the frame size). $G$ is an undirected graph, with the set of edges defined by motion chains (Fig.~\ref{fig:spacetime_graph}).
    
    \textbf{Motion chains} are created by following the optical flow vectors extracted between consecutive frames (function $f$ in Alg.~\ref{alg:cs-gdm} and Fig.~\ref{fig:cs_gdm}). They define paths of motion (optical flow), starting from a pixel and following the flow either forward and backward in time. Thus, multiple chains could pass through a pixel, at least one moving forward and one moving backward. One pixel can have several or no incoming motion chains per direction. Still, there will always be a single outgoing chain for each considered direction (forward or backward). These motion chains directly define our graph structure: two graph nodes (video pixels) $a$ and $b$, nearby or distant in time, are connected by an edge, if and only if they are also connected by at least one motion chain in some direction. Thus, graph edges could be either short or long-range, depending on the temporal length of the motion chain between the two nodes, which always belong to frames at different times. 
        
    The motion chains set up the basic graph structure and also define a \textbf{motion matrix} $\textbf{M} \in \mathbb{R}^{n \times n}$, with non-zero elements corresponding to the edges of the graph (Fig.~\ref{fig:spacetime_graph}, function $g$ in Alg.~\ref{alg:cs-gdm} and Fig.~\ref{fig:cs_gdm}). Each such edge has an associated weight determined by the temporal distance between its corresponding vertices. In consequence, $\textbf{M}_{a,b}=l(a,b) k(a,b)$, where $l:V\times V \rightarrow \{0,1\}$ is a function with $l(a,b)=1$ if $(a,b)\in E$ and zero otherwise, while $k:V\times V \rightarrow \mathbb{R}$ is a Gaussian function modeling the temporal relationship between nodes $a$ and $b$. Thus, $\textbf{M}_{a,b}=k(a,b)$ if $a$ and $b$ are connected and zero otherwise. According to definition, $\textbf{M}$ is symmetric and semi-positive definite, having all non-negative elements. We also expect $\mathbf{M}$ to be very sparse, since the motion chains that define its structure are also sparse (there is only one flow going out in each direction, per pixel).
     
\begin{figure*}[htb]
    \includegraphics[width=1\textwidth]{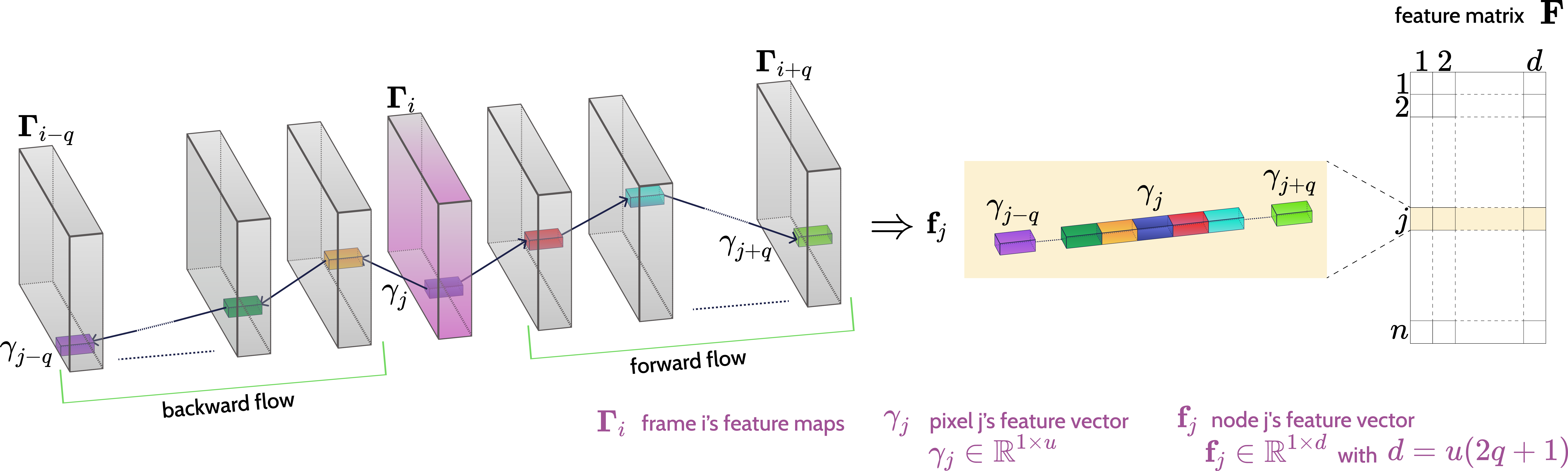}
    \centering
    \caption{Collection of node features along motion chains: for a given node $j$, the features forming feature vector $\mathbf{f}_{j}$ are collected along the two outgoing motion chains (one forward, one backward), from different features of pixels associated to nodes met along the chains. Thus $\mathbf{f}_{j}$ is obtained by concatenating $\gamma_{j+k}$, $k \in [-q,q]$, where $\gamma_{j+k}$ are the features of pixel $j+k$ (met along the chain) and $q$ defines a temporal window. By considering the temporal dimension in constructing node-level features, we capture more distinctive, object-specific local spatio-temporal patterns. The graphs' feature matrix $\mathbf{F}$ is formed by stacking all the node-level feature vectors. [Best viewed in color]}
    \label{fig:chain_features}
    \end{figure*}

    In the space-time graph, each node $a$ has associated \textbf{node-level features} $\textbf{f}_a \in \mathbb{R}^{1 \times d}$. The feature vectors are collected along the outgoing motion chains starting from the pixel associated with node $a$ and passing through all pixels connected to $a$ through motion chains.  In Fig.~\ref{fig:chain_features} we illustrate the process of collecting features along the flow chains (function $h$ in Alg.~\ref{alg:cs-gdm} and Fig.~\ref{fig:cs_gdm}). All node-level feature vectors build the feature matrix $\mathbf{F}\in\mathbb{R}^{n\times d}$. We could consider different types of pixel-level features in practice as we will discuss in detail in Sec.~\ref{sec:exp_analysis_graph_module}. For example, in our fully iterative (cyclic) system, pixel-level features integrate knowledge of the graph-trained Network Module from the previous cycle.
    
     Now we are ready to define the segmentation task as a spectral clustering problem and associate \textbf{node segmentation labels} to each node $a$. We consider (soft) mask labels in the range $[0,1]$ that indicate the node's confidence of being part of the video sequence's primary object. Then, a segmentation solution over the whole spatio-temporal volume can be represented as a vector $\mathbf{x} \in \mathbb{R}^{n \times 1}$. Each element of $\mathbf{x}$ indicates the label of one node and, consequently, the label of its associated pixel. We can also visualize $\mathbf{x}$ as a concatenation of the linearized per-image soft-segmentation masks: $\mathbf{x}=(\mathbf{x}_1, ...\mathbf{x}_m)$, where $\mathbf{x}_i$ is the segmentation map associated to the $i^\text{th}$ frame.
  
    \subsection{Spectral clustering problem formulation}\label{sec:graph_module_problem_formulation}
   
   We consider the main object of interest in video to be the strongest cluster formed by motion chain connections and unary node-level features. On the one hand, nodes belonging to the main object are strongly connected through short and long-range motion chains. On the other hand, object nodes should be predictable from local features to ensure the segmentation map's consistency. Label $x_i$ of node $i$ indicates its association with the object cluster: if the node is part of the object, its label $x_i$ should be 1, while $x_i$ should be 0 if node $i$ is not part of the object. Thus, the object's space-time cluster is defined by the indicator vector $\mathbf{x}$, which contains all object masks for all frames.

    Next, we define the intra-cluster score to be $S=\mathbf{x}^T\mathbf{M}\mathbf{x}$, with the constraint that labels $\mathbf{x}$ should live in the features subspace as explained shortly. We also relax $\mathbf{x}$ to be continuous and impose unit L2-norm. We can enforce the unit norm over the labels vector $\mathbf{x}$ as we are only interested in the computed labels' relative values. Any transformation that preserves this order will not affect the final solution. To solve our segmentation problem, we need to maximize the clustering score $\mathbf{x}^T\mathbf{M}\mathbf{x}$ under the unit norm constraint $\|\mathbf{x}\|_2=1$ and the constraint that $\mathbf{x}$ is a linear combination of the columns of feature matrix $\mathbf{F}$. Note that the linear combination constraint is necessary for obtaining the final spectral solution. It should also be reasonable in practice, especially when the features are powerful and rich. However, as we see in experiments, even simpler motion features work well. Consequently, $\mathbf{x}$ should be predictable from the feature values by linear regression. According to this particular problem setup and the considered constraints, the object pixels will form a strong cluster concerning long-range motion patterns and node-level features. Shortly we will see that solving for the optimal $\mathbf{x}$ becomes finding the principal spectral cluster of a Feature-Motion matrix that couples together motion $\mathbf{M}$ and features $\mathbf{F}$. 
    
	We define matrix $\mathbf{P}$  to be the projection matrix that projects any vector into the columns space of feature matrix $\mathbf{F}$ ($\mathbf{P}=\mathbf{F}(\mathbf{F}^T\mathbf{F})^{-1}\mathbf{F}^T$). The constraint over vector $\mathbf{x}$, which states that it should be a linear combination of the columns of $\mathbf{F}$, can be satisfied by requiring $\mathbf{x}$ to be equal to $\mathbf{P}\mathbf{x}$. The constraint ensures that $\mathbf{x}$ lives in the subspace spanned by the columns of $\mathbf{F}$. Considering the reformulation of the constraints as mentioned above, we obtain the following result:
  
    \begin{proposition} The optimal solution
    $\mathbf{x}^*$ that maximizes $\mathbf{x}^T\mathbf{Mx}$ under constraints $\mathbf{x}=\mathbf{P}\mathbf{x}$ and $\|\mathbf{x}\|_2=1$, will also maximize $\mathbf{x}^T\mathbf{PMPx}$ under constraint $\|\mathbf{x}\|_2$.
    \end{proposition}\label{prop_1}
    \begin{proof}[Proof sketch] As $\mathbf{x}^*$ maximizes $\mathbf{x}^T\mathbf{Mx}$ under constraints $\mathbf{x}=\mathbf{Px}$ and $\|\mathbf{x}\|_2=1$, it also maximizes $\mathbf{(Px)}^T\mathbf{MPx}$. As $\mathbf{P=P^T}$, it follows that $\mathbf{x}^*$ maximizes $\mathbf{x}^T\mathbf{PMPx}$ under the considered constraints. Also note that $\mathbf{P}$ will project any vector onto the subspace spanned by the columns of $\mathbf{F}$. Consequently, the optimal solution of unit norm, $\mathbf{x}^*$, must live in the features subspace. Therefore we conclude that $\mathbf{x}^*=\mathbf{Px}^*$ and the result follows.
    \end{proof}
    
    Considering Proposition 1, the optimization problem can be defined as follows: 
    
    \begin{equation}
        \label{eq_problem1}
        \mathbf{x^*}=\argmax_{\mathbf{x}} \mathbf{x}^T\mathbf{P}\mathbf{M}\mathbf{P}\mathbf{x} \quad \text{s.t.} \quad \|\mathbf{x}\|_2=1
    \end{equation}

    The matrix $\mathbf{PMP}$ is the central element of our graph optimization problem defined in Eq.~\ref{eq_problem1}. We will refer to this matrix as the \textbf{Feature-Motion matrix} and note it with $\mathbf{A}$. $\mathbf{A}$ couples long-range motion patterns incorporated in the motion matrix $\mathbf{M}$ and the projection $\mathbf{P}$ onto the features subspace. Thus, $\mathbf{A}$ represents the final adjacency matrix of our space-time graph. 
    
    \subsection{Graph optimization algorithm}\label{sec:graph_module_optimization_alg}
    The principal eigenvector of the Feature-Motion matrix $\mathbf{A}$ optimally solves the problem defined in Eq.~\ref{eq_problem1}. In this formulation, we transform segmentation into classical spectral clustering~\cite{meila_shi}, also related to spectral approaches from graph matching~\cite{leordeanu2012unsupervised}. Based on the properties of $\mathbf{A}$, having non-negative elements, we can infer, using the Perron-Frobenius theorem, that the optimal solution $\mathbf{x}^*$ has positive values. Our algorithm, which is an efficient implementation of the Power Iteration Method, will converge to the optimal solution $\mathbf{x}^*$. In Algorithm~\ref{alg:practical_iterations}, we introduce the actual steps of our algorithm, which exploits the space-time graph but avoids to compute the $\mathbf{P}\mathbf{M}\mathbf{P}$ matrix, reducing both the computational burden and the memory requirements.
   
    \begin{algorithm}
        \caption{\\main algorithmic steps of \textcolor{m_green}{Graph Module} during iteration t}
        \label{alg:practical_iterations}
        \begin{algorithmic}[1]
            \Statex \textbf{Propagation step}
            \State $\mathbf{x}^{(t)} \gets \mathbf{M}\mathbf{x}^{(t-1)}$ \Comment{$\mathbf{x}^{(t)}$ - node labels from iteration $t$}
            \Statex \textbf{Projection step}
            \State $\mathbf{x}^{(t)} \gets  \mathbf{P}\mathbf{x}^{(t)}$ \Statex \textbf{Normalization step}
            \State $\mathbf{x}^{(t)} \gets \frac{\mathbf{x}^{(t)}}{ \|\mathbf{x}^{(t)}\|_2}$
        \end{algorithmic}
    \end{algorithm}
    
    The \textbf{propagation step} is equivalent to having each node $a$ update its label according to  $x^{(t)}_a=\sum_j\mathbf{M}_{a,b}x^{(t-1)}_b$. The step is also equivalent to each node $a$ having its own label being propagated to all nodes it is connected to. To efficiently implement this, we start from a given node $a$ and cast its information to all the nodes $b$ met along the motion chains, considering both forward and backward directions. When passing through a node $b$, we update its label as $x_b \gets x_b + \mathbf{M}_{a,b}x_a$ but also update the label of $a$ accordingly $x_a \gets x_a + \mathbf{M}_{a,b}x_b$. Thus, we jointly propagate information from all the nodes in one frame to all neighboring frames in both forward and backward directions. The votes are cast only to frames within a radius of $p$ time steps from the reference frame as the pairwise terms $\mathbf{M}_{a,b}$ rapidly decrease towards zero with the temporal distance between nodes $a$ and $b$. 
    This particular implementation significantly reduces the computational burden (as shown in Sec.~\ref{sec:graph_module_comp_complexity}), performing power iteration on the Feature Motion matrix $\mathbf{A}$  and efficiently computing its leading eigenvector without having to build the huge $\mathbf{A}$
    
    The role of the \textbf{projection step} is to force the solution $\mathbf{x}$ to live in the column space of $\mathbf{F}$. To ensure this constraint, we require $\mathbf{x}$ to be predictable from $\mathbf{F}$ by ridge regression. During each iteration, we estimate the optimal set of weights $\mathbf{w}^*$ that best approximate current node labels given node-level features in $\mathbf{F}$. The weights are computed as $\mathbf{w}^* = (\mathbf{F^TF})^{(-1)}\mathbf{F}^T\mathbf{x}^{(t)}$, considering current labels estimates $\mathbf{x}^{(t)}$. Further, the node labels are updates by linear projection on the features subspace: $\mathbf{x}^{(t)}\gets\mathbf{F}\mathbf{w}^* = \mathbf{P}\mathbf{x}^{(t)}$. 

    During the \textbf{normalization step}, the labels vector is divided by its norm to ensure the unit L2-norm constraint. We highlight that this normalization does not affect the relative order of the elements of $\mathbf{x}$; thus, it is not modifying the final solution of our segmentation problem.
    
     \begin{figure*}[h]
    \includegraphics[width=1\textwidth]{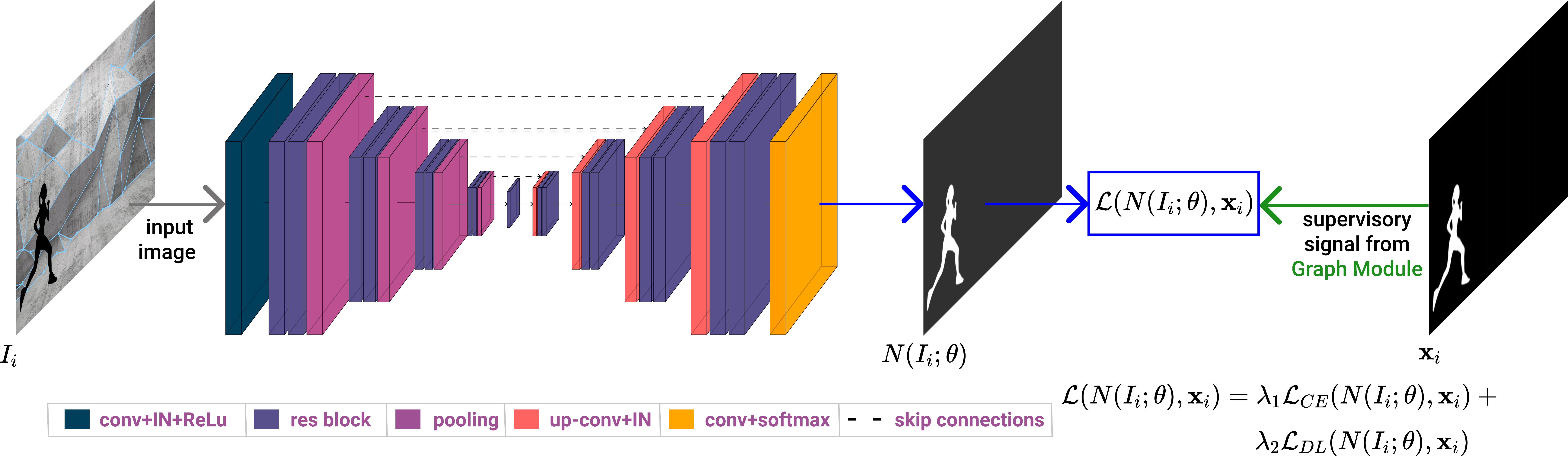}
    \centering
    \caption{The Network Module. We use a UNet-like architecture with dilated convolutions and residual blocks.  Our network's input is a RGB image, while the output is the segmentation mask associated with the primary object of the video sequence. The network is trained considering the pseudo ground truth provided by the Graph Module. Further, the learned representation will reinitialize the graph optimization problem in a new cycle. The presented segmentation maps are only illustrative. [Best viewed in color]}
    \label{fig:DoS}
    \end{figure*}
    
    \subsection{Theoretical analysis}\label{sec:graph_module_th_analysis}
	  
	Our power iteration algorithm efficiently solves the clustering problem introduced in Eq.~\ref{eq_problem1}, which defines the segmentation task over space and time. The main trick is to avoid explicitly representing the adjacency matrix $\mathbf{M}$ and the Feature-Motion matrix $\mathbf{A}$. In the current section, we analyze the algorithm and prove that it converges to the optimal solution of the optimization problem Eq.~\ref{eq_problem1}.
	
    \begin{proposition}
    Algorithm~\ref{alg:practical_iterations}  converges to the principal eigenvector of the Feature-Motion matrix $\mathbf{A=PMP}$, which is the global optimum of the optimization problem defined in Eq.~\ref{eq_problem1}.
    \end{proposition}\label{prop_2}
    
    \begin{proof}[Proof Sketch]
    If we try to formulate the steps of our algorithm into a single update, we lead to the recurrence relation describing the power iteration: $\mathbf{x}^{(t)}=(\mathbf{PM}\mathbf{x}^{(t-1)})/(\|\mathbf{PM}\mathbf{x}^{(t-1)}\|_2)$. This means that the proposed algorithm is guaranteed to converge to the principal eigenvector of the $\mathbf{PM}$ matrix. It follows that $\mathbf{x^*}$ maximizes the Rayleigh quotient $R(\mathbf{PM}, \mathbf{x})=(\mathbf{x^TPMx})/(\mathbf{x^Tx})$. From the projection and normalization steps of Algorithm~\ref{alg:practical_iterations}, it results that the L2-norm of our optimal solution is 1 ($\|\mathbf{x^*}\|_2=1$) and $\mathbf{x}^*$ lives in the column space of $\mathbf{F}$, meaning that $\mathbf{x^*}=\mathbf{Px^*}$. It immediately results that the optimal solution $\mathbf{x^*}$ also maximizes our objective  $\mathbf{x^TPMPx}$, under unit L2-norm constraints. According to this result, the convergence point of our algorithm is the global optimum of graph optimization problem defined in Eq.~\ref{eq_problem1}, which corresponds to the leading eigenvector of the Feature-Motion matrix $\mathbf{A}=\mathbf{PMP}$.
    \end{proof}
   
     We showed that the proposed algorithm is guaranteed to converge to our clustering problem's global optimum. Thus our method always identifies the primary object in the video as the strongest cluster of the defined space-time graph, in terms of appearance and motion, which should intuitively correspond to the most noticeable object. 

 \subsection{Computational complexity}\label{sec:graph_module_comp_complexity}

  Given the formulation, with one-to-one correspondences between video pixels and graph nodes, the problem of spectral clustering in the whole graph could appear to be intractable. Nevertheless, the proposed algorithm computes the spectral solution efficiently by avoiding to build the graph adjacency matrix explicitly. The \textbf{propagation step} has complexity $O(np)$, as we propagate labels of one node on a radius of $p$ time steps, considering that $\mathbf{M}_{a,b}=k(a,b)$ decreases rapidly towards zero with the temporal distance between nodes $a$ and $b$. The \textbf{projection step} has complexity $O(d^2n)$, as the most expensive step is the matrix multiplication $\mathbf{F}^{T}\mathbf{F}$. We highlight that in all the considered setups, $d<100$. The projection step's complexity asymptotically dominates all other steps, resulting in an overall complexity $O(d^2n)$ for one iteration. In practice, the algorithm can be further optimized by precomputing elements, such as the projection matrix $\mathbf{P}$, and avoid redundant computations during iterations. In order to compute optical flow information, FlowNet2.0 requires $0.04$ sec/frame, while RAFT requires $0.33$ sec/frame. In our unsupervised implementation, we take $0.31$ sec/frame to precompute information necessary for the propagation and regression steps and $0.53$ sec/frame for the optimization process. For the supervised scenario, we also need to consider the time required for the forward pass through our backbones, which is less than $0.1$ sec/frame. The runtimes reported above correspond to our implementation in Python with PyTorch on a computer with the following specifications: Intel(R) Xeon(R) CPU E5-2697A v4 @ 2.60GHz, GPU GeForce GTX 1080.   
    
\section{Network Module}\label{sec:network_module}
    
The Network Module is a deep segmentation model that complements the space-time graph. At each cycle, the network is trained from scratch solely using as supervisory signal the output of the Graph Module. The graph, by taking advantage of the temporal dimension, discovers object properties, which the network, with its higher representation power, could learn in single frames. The network indeed learns high-level representations of the main object and passes them to the graph at the following iteration of clustering. These newly learned deep features are added to the feature matrix $\mathbf{F}$ in order to re-initialize the space-time graph for a new cycle. In general, we observed that the graph always has an advantage over the net due to the temporal dimension, such that, at a given cycle, the network does not quite match the graph's performance. However, the features learned by the network from the output of the graph always help the graph at the next cycle, and the overall system gains a significant boost.
 
    
 The Network Module is trained on pairs of samples $(I_i, \mathbf{x}_i)$, where $I_i\in\mathbb{R}^{h\times w\times 3}$ is the $i^{\text{th}}$ image of the video sequence, and $\mathbf{x}_i\in [0,1]^{h\times w}$ is the supervisory signal, for frame $i$, provided by the Graph Module. The loss function is defined as a weighted average between cross-entropy loss and dice loss and is computed as:
  
    \begin{equation}
    \begin{aligned}
     \mathcal{L}(N(I_i; \theta), \mathbf{x}_i) = & \lambda_1\mathcal{L}_{CE}(N(I_i; \theta), \mathbf{x}_i)+\\
     & \lambda_2\mathcal{L}_{DL}(N(I_i; \theta), \mathbf{x}_i)
    \end{aligned}
    \end{equation}
    
   where $N(I_i;\theta)$ represents the network's prediction for input frame $I_i$ considering network parameters $\theta$. For cross entropy loss we binarize our pseudo ground truth while we employ the soft version of the dice loss. This configuration ensures an increase penalty in high confidence areas, while we ensure a more permissive behaviour in areas of uncertainty. In practice, we consider $\lambda_1=\lambda_2=0.5$. The Network Module solves the following optimization task:
   \begin{equation}\label{eq:net_optim}
    \begin{aligned}
    \theta^{*} =\argmin_{\theta}\frac{1}{m}\sum^{m}_{i=1}\mathcal{L}(N(I_i; \theta), \mathbf{x}_i)
    \end{aligned}
   \end{equation}
  
    Minimizing the objective function of Eq.~\ref{eq:net_optim}, leads to the optimal network parameters $\theta^*$, associated to current cycle of our self-supervised system. Our networks predictions $\mathbf{s}_i=N(I_i;\theta^*)$ will be considered as pixel-level features in the next cycle. Thus, the Graph Module will benefit from the high-level representation inferred by the Network Module, overcoming its initial limitation. 

    \begin{figure*}[htb]
        \includegraphics[width=1\textwidth]{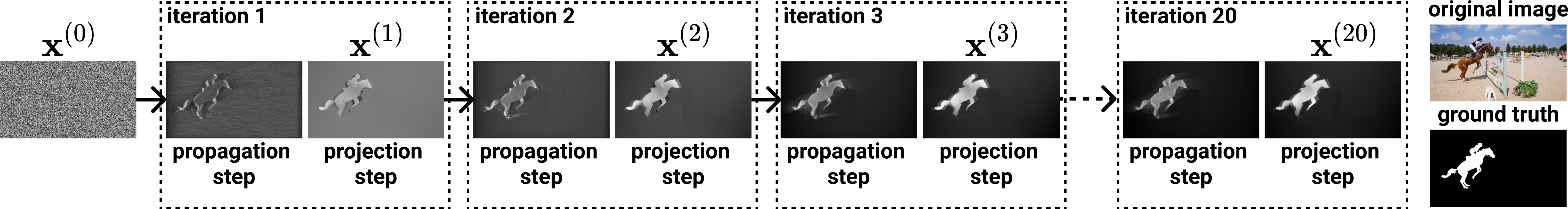}
        \centering
        \caption{Evolution of segmentation solution $\mathbf{x}$ over several iterations of our graph optimization algorithm (Alg.~\ref{alg:practical_iterations}). We present the evolution of both propagation and projection steps (followed by normalization), considering random initial node labels $\mathbf{x}^{(0)}$.}
        \label{fig:convergence}
    \end{figure*}

\section{Datasets and Metrics}\label{sec:exp_setup}
    
    \textbf{Datasets:} for the experimental analysis, we chose three popular video object segmentation datasets (DAVIS2016~\cite{Perazzi2016}, SegTrackv2~\cite{li2013video} and YouTube-Objects~\cite{prest2012learning}) and one recently published video salient object detection dataset: (DAVSOD~\cite{fan2019shifting}). All considered datasets have pixel-level annotations of the target object. DAVIS2016 contains 20 high-resolution video sequences, with drastic appearance changes, occlusions, and motion blur. Segtrackv2 is composed of 14 challenging video sequences, containing objects that are not usually present in annotated datasets (e.g., worm). \cite{prest2012learning} initially published YouTube-Objects dataset with bounding box annotations, and we use a subset comprising of 126 video sequences ($\approx26k$ frames) with pixel-level annotations, introduced in \cite{jain2014supervoxel}. DAVSOD contains 226 video sequences ($\approx24k$ frames) with realistic scenes. We use the 'Easy' test set, for which a benchmark is established in \cite{fan2019shifting}. We have chosen this dataset as the salient object detection task's video version is strongly related to our task. We highlight that for all considered datasets, the main object can be represented by one or a group of several tightly connected objects behaving like one, such as a man riding a bicycle. DAVIS contains only high-resolution video sequences, while the other datasets contain videos at a relatively low resolution. While DAVIS comprises shorter and very dynamic sequences, the other datasets have longer videos with complex scenes containing dynamic and static subsequences.   \textbf{Metrics:} following the established methodology, we report the average intersection over union score (J Mean) for all video object segmentation datasets, along with boundary  F-Measure (F Mean) for DAVIS. For DAVSOD, we report F-Measure (F) \cite{achanta2009frequency} and mean absolute error (MAE). When necessary, we will also report relative percentage changes of our method concerning various baselines. Considering two methods with performances $v_1$ and $v_2$, we define the relative percentage change of second method w.r.t. to the first, as $100 * (v_2-v_1)/v_1$. \textbf{Resolution:} in all our experiments, our system works at a resolution of $416 \times 224$ pixels. 

    \begin{figure}[htb]
    \centering
    \includegraphics[width=1\columnwidth]{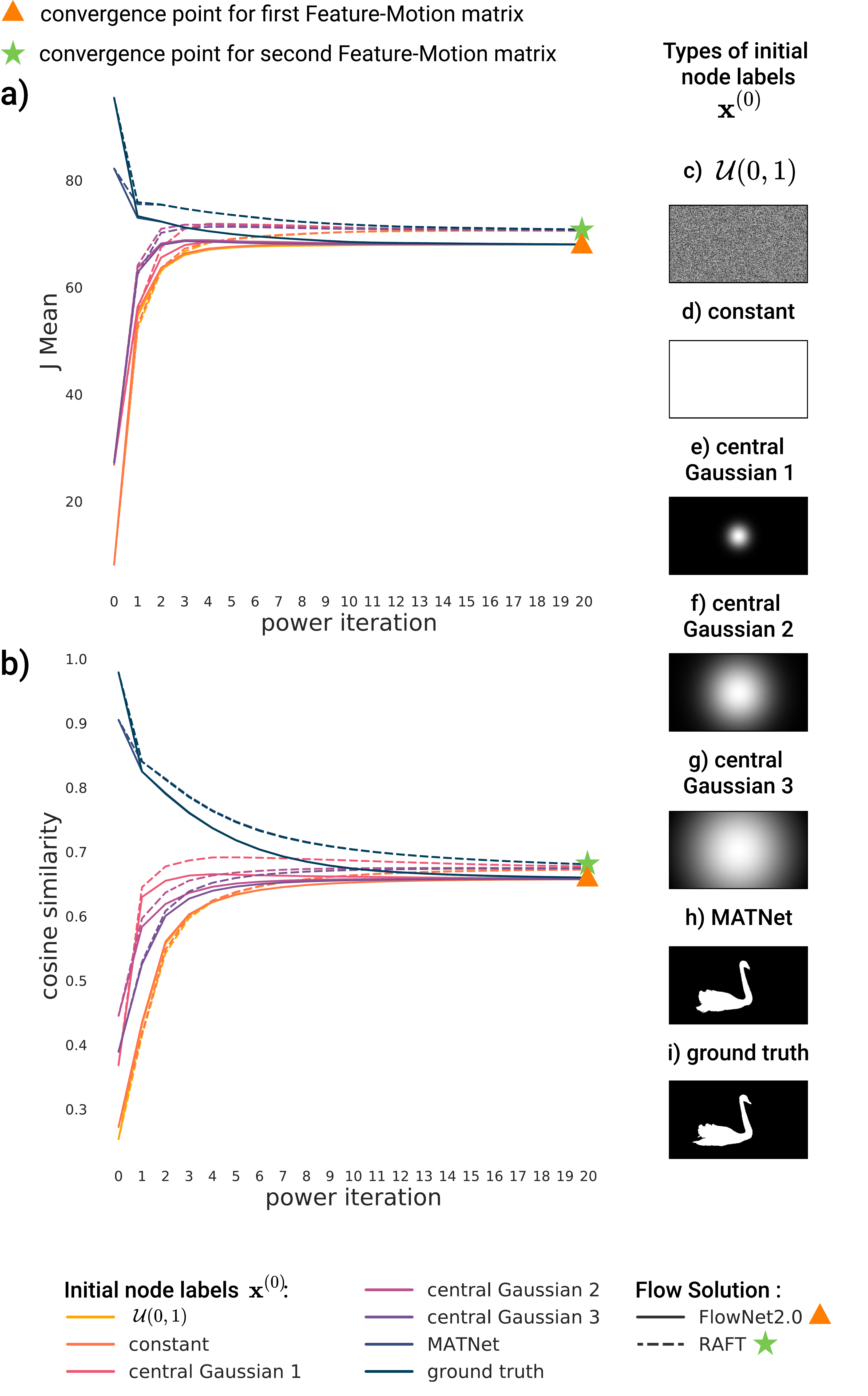}
    \caption{Convergence analysis of the Graph Module considering different initial node labels $\mathbf{x}^{(0)}$ and two optical flow methods. The flow solutions will define two different Feature-Motion matrices, and in consequence, different converge points. a) Evolution of J Mean b) Evolution of cosine similarity between $\mathbf{x}$ and ground truth c)-i) visual representation of initial node labels $\mathbf{x}^{(0)}$. Experiments were conducted on full DAVIS2016  dataset. [Best viewed in color]} 
    \label{fig:convergence_analysis}
    \end{figure}
     
\section{Experimental Analysis of the Graph} 
\label{sec:exp_analysis_graph_module}

    We begin with a numerical and experimental analysis of the first graph module (first cycle), which represents the system's main unsupervised teacher.  We are interested in several aspects. First, we want to verify that the graph optimization algorithm indeed converges in practice to a unique, global solution, namely the Feature-Motion matrix's principal eigenvector (Sec. \ref{sec:exp_analysis_graph_module_conv}). Next, we want to understand better the influence of the optical flow module on performance. How sensitive is the accuracy of space-time clustering to the accuracy of the optical flow. In Sec.~\ref{sec:exp_analysis_graph_module_flow}, we show that the graph optimization algorithm does require a good quality optical flow, but it is also robust to the choice of the actual flow method used. Even when the flow chosen is not state-of-the-art, performance drops only slightly. While our main focus is unsupervised learning, we also want to show that our method is general and can easily incorporate supervised information. Thus, in the next Sec.~\ref{sec:exp_analysis_graph_module_features}, we look more closely at how we could integrate supervised features in the feature matrix $\mathbf{F}$ and further boost performance. The addition of such supervision signal is crucial in competing with top supervised methods on various datasets, as shown in the experimental  section~\ref{sec:exp_analysis_quant_qual_comparison}. We conclude our analysis of the graph module with a detailed, in-depth look over the Feature-Motion matrix's spectral structure (Sec.~\ref{sec:exp_analysis_graph_module_fm}). We show that it indeed contains a strong natural cluster in practice, which is reflected on the principal eigenvector and corresponds to the main object in the sequence.
    
\subsection{Convergence of optimization algorithm in practice}\label{sec:exp_analysis_graph_module_conv}
    In the theoretical section~\ref{sec:graph_module_th_analysis}, we showed that the graph algorithm (Alg.~\ref{alg:practical_iterations}) is an iterative process equivalent to the classical power iteration. It computes the principal eigenvector of the Feature-Motion matrix $\mathbf{A}$, which is the unique, global solution of the relaxed segmentation problem in  Eq.~\ref{eq_problem1}. This implies that the segmentation process should converge to the same solution $\mathbf{x}^*$ regardless of its initialization $\mathbf{x}^{(0)}$. This section verifies that this property indeed holds in practice and that the algorithm converges to the main object in the video even when the initial starting solution is completely random. As mentioned before, in the last subsection (\ref{sec:exp_analysis_graph_module_fm}), we also verify that the Feature-Motion matrix has a main strong cluster, which, as the experiments show, does correspond to the main object in the sequence, according to the humanly labeled ground truth.
    
    In Fig.~\ref{fig:convergence}, we display a representative example, showing the evolution of the soft-segmentation masks (the optimization solution) over several iterations of our algorithm. We start with uniform distributed labels for our nodes (random initial solution). With each iteration of  Alg.~\ref{alg:practical_iterations} we see how the main object emerges naturally, and its mask becomes more and more precise. This means that the elements of the eigenvector solution $\mathbf{x}^{(t)}$ corresponding to the object are becoming stronger and stronger with each iteration $t$, while the others get closer to zero until a final convergence point $\mathbf{x}^*$ is reached.
    
    To verify the convergence property to a unique solution, we take a closer look at the starting point's influence in practice. Thus, we verified what happens when given the same Feature-Motion matrix (which depends only on the optical flow module used and not on the initial solution $\mathbf{x}^{(0)}$), we change the initial starting points. We considered vastly different initial solutions, such as some that are completely ignorant of the actual object in the image (uniform distribution, constant, central Gaussians with different standard deviations) and others that are already close to the ground truth (produced by MATNet~\cite{zhou2020matnet}, a top published method as well as the ground truth itself). We also consider two different Feature-Motion matrices, corresponding to two different optical flow solutions: FlowNet2.0~\cite{ilg2017flownet} and RAFT~\cite{teed2020raft}. We performed all tests on the DAVIS dataset, over 20 iterations, and present the results in Fig.~\ref{fig:convergence_analysis}. The main interesting observation is that, indeed, our results confirm the theoretical expectations very well. Our algorithm reaches the same good quality solution, regardless of initialization, which makes it robust to the lack of prior knowledge in the unsupervised case. Simultaneously, as expected, the final solution depends only on the Feature-Motion matrix $\mathbf{A}$, as it should be, since $\mathbf{A}$ has all the relevant information about how that object looks and moves. We also note that the speed of convergence depends on the initial solution: the closer the initial point is to the final eigenvector, the faster the convergence.
    
\subsection{Unsupervised case: influence of optical flow}
\label{sec:exp_analysis_graph_module_flow}

    The optical flow is the basis for establishing the short and long-range connections in the graph through the motion chains defined in Sec.~\ref{sec:graph_module_spacetime_graph}. Thus, two nodes (pixels) connected in the motion chain are also connected in the graph, whereas nodes that are not connected by motion chains are not connected in the graph either. The connectivity is encoded in matrix $\mathbf{M}$ and immediately transfers to the Feature-Motion matrix $\mathbf{A}$, which is the adjacency matrix of the space-time graph. However, optical flow also influences the feature matrix $\mathbf{F}$ since we construct node-level features using optical flow.
   
    To compute the node-level features, we follow the procedure described in Sec.~\ref{sec:graph_module_spacetime_graph} and collect features along the outgoing motion chains. In this case, the features are the optical flow displacement 2D vectors, from one frame to the next, centered around a given point backward and forward in time. Such features encode each point's motion pattern, the assumption being that points with common patterns tend to belong to the same object and that the main object has a set of distinctive motion patterns. 
    
    Therefore, optical flow determines both the motion matrix $\mathbf{M}$ and feature matrix $\mathbf{F}$ and is at the core of the overall Feature-Motion matrix $\mathbf{A}$. It is expected that the accuracy of the flow will influence the quality of the results. The real question is by how much the flow affects the final result and how robust our algorithm is to errors in flow.
   
    To ensure that our approach is indeed unsupervised, in the sense that it uses no manual annotations, we used optical flow methods (FlowNet2.0~\cite{ilg2017flownet} and RAFT~\cite{teed2020raft}) that are trained without any human supervision on pure synthetic datasets: FlowNet2.0 is trained on FlyingThings3D~\cite{mayer2016large} and FlyingChairs~\cite{dosovitskiy2015flownet}, while RAFT is trained on FlyingThings3D~\cite{mayer2016large}. Note that since optical flow is a low-to-mid level task (its precise goal being to estimate the well-defined motion field), training on purely synthetic data is possible, with great success, unlike the higher-level task of object-level segmentation. The two methods have different performance levels as shown in their ranking on MPISintel \cite{butler2012naturalistic, wulff2012lessons} leaderboard, which is the main benchmark of optical flow methods: RAFT ranks first (and third on KITTI-2015 \cite{Menze2015CVPR}), while FlowNet2.0 occupies only the $133^\text{rd}$ position on MPISintel. We perform comparative tests, using both optical flow approaches, again on DAVIS dataset.

    Even though the two optical flow approaches are of different levels of performance, as shown in our tests (Table~\ref{tab:exp_analysis_features_unsup}), the algorithm is not too sensitive to the choice of flow, with only a $2.6\%$ gap in terms of J Mean between the two configurations. While a certain degree of sensitivity is expected, our graph optimization can still maintain a good performance level, even in this purely unsupervised scenario that fully depends on flow. The relatively small difference between the solutions obtained with one type of flow vs. the other could also be well visualized in Fig.~\ref{fig:convergence_analysis}, where it can be easily seen that the two points of convergence (one per optical flow method) are almost equally close to the ground truth.
   
    In Table~\ref{tab:exp_analysis_features_unsup} we also present a different experiment in which for a given optical flow used to build the graph motion structure ($\mathbf{M}$), we concatenated node-level feature vectors computed with both optical flow methods (RAFT and FlowNet2.0) to build $\mathbf{F}$. Note how combining the two sources of information significantly boosts the performance over each of the two choices, indicating that the different optical flow approaches could contain some independent, complementary information and that the node-level features' role is very important for good accuracy. This observation leads to the next set of experiments in which we further enhance the node-level features in $\mathbf{F}$ with supervised information.

    \begingroup
    \setlength{\tabcolsep}{1pt}
    \begin{table}
        \centering
        \captionsetup{justification=centering}
        \caption{\\Performance of Unsupervised Graph Module ($1^{st}$ cycle)}\label{tab:exp_analysis_features_unsup}
        \begin{tabular}{cccccc} 
        \toprule
        \multicolumn{6}{l}{\textcolor{m_green}{Unsupervised method}} \\
        \toprule
         & $\mathbf{M}$ & $\mathbf{F}$ & J Mean $\uparrow$ & F Mean $\uparrow$ & Avg $\uparrow$ \\
        \cmidrule(lr){2-3}
        \cmidrule(lr){4-6}
        \multirow{5}{*}{{\shortstack{\textcolor{m_green}{Graph}\\\textcolor{m_green}{Module} }}} & \multirow{2}{*}{FlowNet2.0~\cite{ilg2017flownet}} & FlowNet2.0 & 68.1 & 65.6 & 66.9 \\
        \cmidrule{3-6}
        & & \shortstack{FlowNet2.0; RAFT} & 70.6 & 68.4 & 69.5 \\
        \cmidrule{2-6}
        & \multirow{2}{*}{RAFT~\cite{teed2020raft}} & RAFT & 70.7 & 67.0 & 68.9 \\
        \cmidrule{3-6}
        & & \shortstack{FlowNet2.0; RAFT} & 70.8 & 68.6 & 69.7 \\
        \bottomrule
    \end{tabular}
     \parbox{\columnwidth}{\footnotesize%
\vspace{1eX}\textit{Tests are performed on the full DAVIS dataset, considering different flow methods. There is a relatively small performance gap between the configuration using FlowNet2.0 for the graph structure and the one using RAFT, which is caused by the gap in ranking and performance between the two flow methods. When the feature matrix of the RAFT configuration is enriched with features of FlowNet2.0, we observe a performance boost, showing that the addition of potentially complementary signals to the node-level features improves the performance of the graph.}}
    \end{table}
    \endgroup

\subsection{Adding supervision to node-level features}
\label{sec:exp_analysis_graph_module_features}

    To better understand the influence of the node-level features used in $\mathbf{F}$, we now consider the case of adding more powerful features which indirectly contain human supervision through various supervised pretraining scenarios. Adding supervised information is also very useful to show that our approach is not limited to the first unsupervised case. To create such supervised node-level features (to be introduced in $\mathbf{F}$) we first consider two deep network backbone models: FCN \cite{long2015fully} with ResNet101 \cite{he2016deep} and DeepLabv3 \cite{chen2017rethinking} with ResNet101. Both models are pretrained, with human supervision, on the COCO dataset \cite{lin2014microsoft} for the related semantic segmentation task. Note that the supervised information only influences $\mathbf{F}$, while the initial node labels ($\mathbf{x}^{(0)}$) are in all our comparative experiments shown next, completely uninformed (drawn from a uniform distribution). We also perform tests when supervised features are taken from other video segmentation approaches. The two scenarios: with image-level features from vanilla backbones or with stronger features from top video segmentation systems, are considered for different experiments, as shown in the remaining of the paper. The actual features are the soft-values collected from the vanilla backbones' last output layer or the different published video segmentation systems. 
    
    These node-level features are collected along the outgoing motion chains, as previously illustrated in Fig.~\ref{fig:chain_features}, the same way as in the unsupervised case when only optical flow vectors were considered. The process is similar to a 3D convolution guided by optical flow information, in the sense that neighbors in time are established considering flow displacements rather than mere location information. The feature collection mechanism helps the feature projection step to ensure spatio-temporal consistency, such that node-level features along the chain are expected to belong to the same physical point on the object. As also illustrated in Fig.~\ref{fig:chain_features}, features are collected considering a temporal window of size $2q+1$, further referred to as \textit{feature chain size}. We will show shortly how this temporal window affects performance. Before that, we focus on the effect of adding such supervised features.
    
    The first experiments presented, performed on DAVIS dataset (Table~\ref{tab:exp_analysis_features_sup}), clearly show that the addition of more powerful features significantly improves the performance of the graph algorithm. Also, note that our space-time graph optimization only, with no graph-network modules cycles, improves over the initial backbones (used to provide node-level features) by more than $10\%$ in absolute J Mean (lines 1-6). The main reason behind this very strong performance boost lies in the core properties of our space-time graph optimization. It considers how the object moves in time and discovers the object as a single cluster in both space and time, versus the pretrained backbones, which process information only at the single-frame level.
    
    \begingroup
    \setlength{\tabcolsep}{1pt} 
    \begin{table}
        \centering
        \captionsetup{justification=centering}
        \caption{\\Performance of Supervised Graph Module ($1^{st}$ cycle)  }\label{tab:exp_analysis_features_sup}
        \begin{tabular}{clcccc}
            \toprule
            \multicolumn{6}{l}{\textcolor{red}{Supervised method}}\\ 
            \toprule
            & & \shortstack{Optical flow} & J Mean $\uparrow$ & F Mean  $\uparrow$ & Avg  $\uparrow$ \\
            \midrule
            \multirow{6}{*}{\rotatebox[origin=c]{90}{\parbox{1.5cm}{\shortstack{image-level\\supervision}}}} 
            & \multicolumn{1}{l}{FCN backbone} & -  & 66.5 & 64.3 & 65.4\\
            \cmidrule{2-6}
            & \multirow{2}{*}{\shortstack{\textcolor{m_green}{Graph  Module}\\ with FCN}} & \multirow{1}{*}{FlowNet2.0~\cite{ilg2017flownet}} & \shortstack{76.6} & \shortstack{74.6} & \shortstack{75.6} \\
            & & \multirow{1}{*}{RAFT~\cite{teed2020raft}} & \shortstack{78.2} &  \shortstack{75.9} & \shortstack{ 77.1}\\
            \cmidrule{2-6}
            & \multicolumn{1}{l}{DeepLabv3 backbone} & -& 67.0 & 64.8 & 65.9 \\
            \cmidrule{2-6}
            & \multirow{2}{*}{\shortstack{\textcolor{m_green}{Graph  Module}\\ with DeepLabv3}} & \multirow{1}{*}{FlowNet2.0~\cite{ilg2017flownet}} & \shortstack{77.4} & \shortstack{74.7 } & \shortstack{76.1} \\
            & & \multirow{1}{*}{RAFT~\cite{teed2020raft}} & \shortstack{77.7} & \shortstack{75.3} & \shortstack{
             76.5} \\
            \midrule
            \multirow{12}{*}{\rotatebox[origin=c]{90}{\parbox{4cm}{\centering\shortstack{video-level\\supervision}}}} & \multicolumn{1}{l}{3DC-Seg~\cite{mahadevan2020making}} & -& 84.3 & 84.7 & 84.5  \\
            \cmidrule{2-6}  
            &\multirow{2}{*}{\shortstack{\textcolor{m_green}{Graph  Module}\\ with 3DC-Seg}} & \multirow{1}{*}{FlowNet2.0~\cite{ilg2017flownet}} & 85.5 & 84.7 & 85.1 \\
            & & \multirow{1}{*}{RAFT~\cite{teed2020raft}} & 85.9 & 85.3 & 85.6 \\
           \cmidrule{2-6}  
            & \multicolumn{1}{l}{MATNet~\cite{zhou2020matnet}} & -& 82.4 & 80.7 & 81.6  \\
            \cmidrule{2-6}  
            &\multirow{2}{*}{\shortstack{\textcolor{m_green}{Graph  Module}\\ with MATNet}} & \multirow{1}{*}{FlowNet2.0~\cite{ilg2017flownet}} & \shortstack{84.1} & \shortstack{82.0} & \shortstack{83.1} \\
            & & \multirow{1}{*}{RAFT~\cite{teed2020raft}} & \shortstack{84.5} & \shortstack{82.4 } & \shortstack{83.5}  \\
           \cmidrule{2-6}  
            & \multicolumn{1}{l}{AnDiff~\cite{yang2019anchor}} & - & 81.7 & 80.5 & 81.1 \\
            \cmidrule{2-6}
            & \multirow{2}{*}{\shortstack{\textcolor{m_green}{Graph  Module}\\ with AnDiff}} & \multirow{1}{*}{FlowNet2.0~\cite{ilg2017flownet}} & \shortstack{82.9 } & \shortstack{82.0} & \shortstack{82.5}\\
            & & \multirow{1}{*}{RAFT~\cite{teed2020raft}} & \shortstack{83.7} & \shortstack{83.1 } & \shortstack{83.4}  \\
            \cmidrule{2-6} 
            & \multicolumn{1}{l}{COSNet~\cite{lu2019see}} & - & 80.5 & 79.5 & 80.0 \\
            \cmidrule{2-6}
            & \multirow{2}{*}{\shortstack{\textcolor{m_green}{Graph  Module}\\ with COSNet}} & \multirow{1}{*}{FlowNet2.0~\cite{ilg2017flownet}} & \shortstack{82.8 } & \shortstack{81.7 } & \shortstack{82.3 }  \\
            & & \multirow{1}{*}{RAFT~\cite{teed2020raft}} & \shortstack{83.4} & \shortstack{81.7} & \shortstack{82.6}   \\
            \bottomrule
        \end{tabular}
         \parbox{\columnwidth}{\footnotesize%
\vspace{1eX}\textit{Experiments are performed on the full DAVIS dataset, considering different supervised features and two optical flow solutions. Features are extracted from semantic segmentation solutions (image-level) and video object segmentation solutions (video-level). We also report the performance of the considered baselines. We emphasize the performance boost brought by our graph optimization algorithm over the baseline methods. }}
    \end{table}
    \endgroup
    
    \begin{figure}
    \includegraphics[width=1\columnwidth]{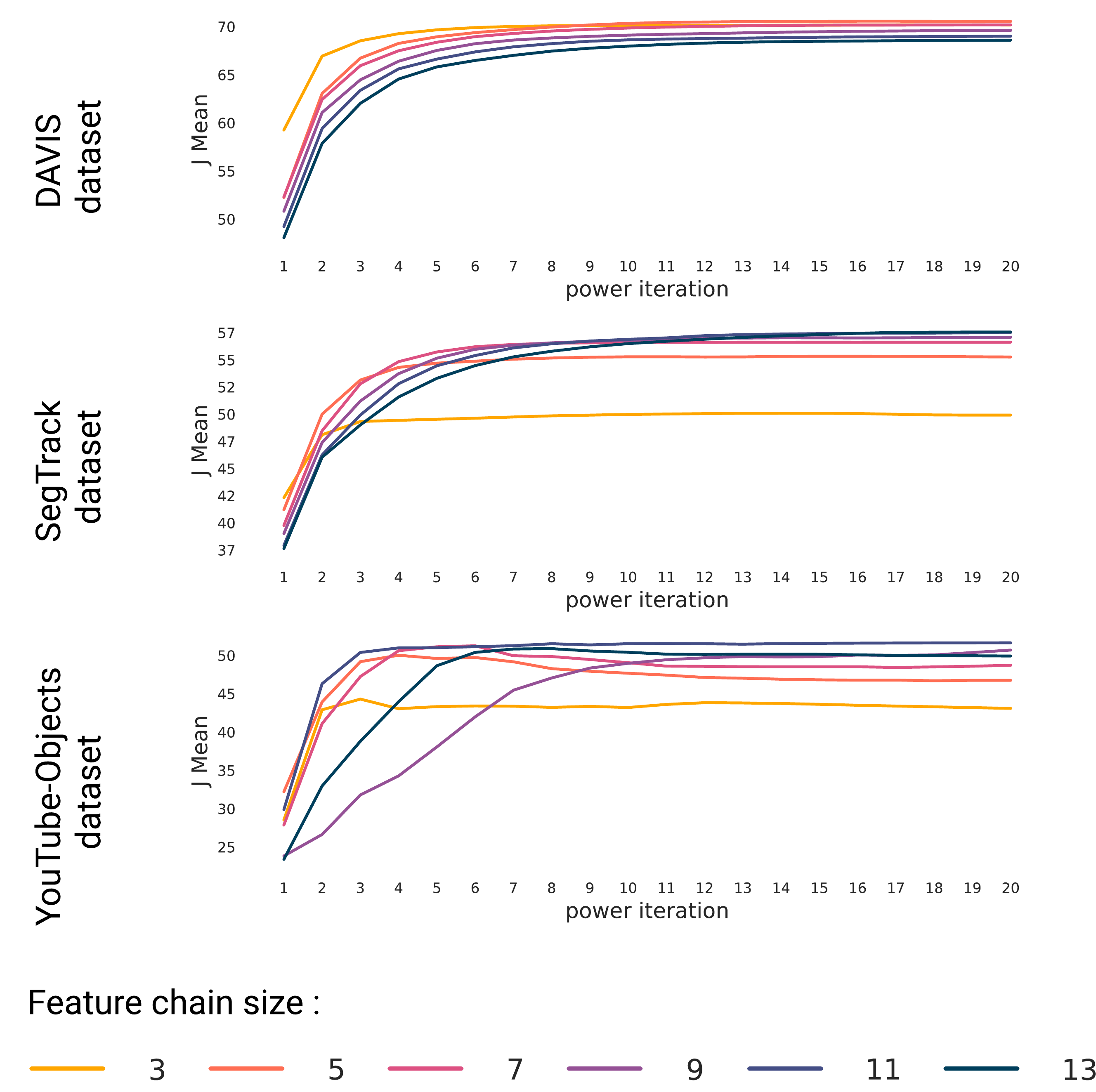}
    \centering
    \caption{Evolution of J Mean for different node feature chain sizes. The experiment was conducted considering our graph optimization algorithm's unsupervised formulation on three datasets: DAVIS2016, YouTube-Objects, and SegTrackv2. In general, longer chains are preferred. [Best viewed in color]} 
    \label{fig:feat_chain_size}
    \end{figure} 
    
    The next experiments shown in the remaining of Table~\ref{tab:exp_analysis_features_sup}, prove that our space-time clustering algorithm alone can bring notable improvements even over the state-of-the-art outputs used for providing node-level features. The top methods considered are 3DC-Seg~\cite{mahadevan2020making}, MATNet~\cite{zhou2020matnet}, AnDiff~\cite{yang2019anchor} and COSNet~\cite{lu2019see}. As expected, the margin of improvement is much smaller, making perfect sense considering that the state-of-the-art methods used as sources of our features already integrated sophisticated ways of considering the temporal dimension. However, the good news is that our approach still manages to bring some valuable information by space-time spectral clustering. It is worth mentioning that in all the experiments, our approach was able to improve over the baseline that provided the node-level features. Therefore, from this perspective, our approach could also be seen as a general segmentation refinement procedure that could be combined with virtually any other segmentation algorithm.
    
    Now, we switch attention to the more technical aspect of feature chain sizes, which establishes for what temporal window around a given point we should consider these features. On the one hand, the longer the size, the more information we have about that particular point on the object as it changes back and forth in time around the current node. On the other hand, the longer the chain size, the higher the likelihood of drift due to optical flow errors.  In Fig.~\ref{fig:feat_chain_size}, we present the performance evolution for different chain sizes in the unsupervised case (when only flow displacements are considered as node features). We test on three datasets, namely DAVIS, YouTube-Objects, and SegTrack. We can make some interesting observations: DAVIS has shorter and more dynamic video sequences, while YouTube-Objects and SegTrack tend to contain longer sequences with more static sub-sequences. Consequently, for DAVIS, smaller chain sizes of $3$ temporal units are preferred, while for YouTube-Objects and SegTrack, chain sizes of up to $13$ temporal units are optimal. However, we also notice that on all three sets, longer chain sizes are always close to the top performers, which again suggests that the time dimension is important when gathering node-level information.
   
\subsection{Spectral analysis of the Feature-Motion matrix}\label{sec:exp_analysis_graph_module_fm}

\begin{figure}[htb]
    \includegraphics[width=1\columnwidth]{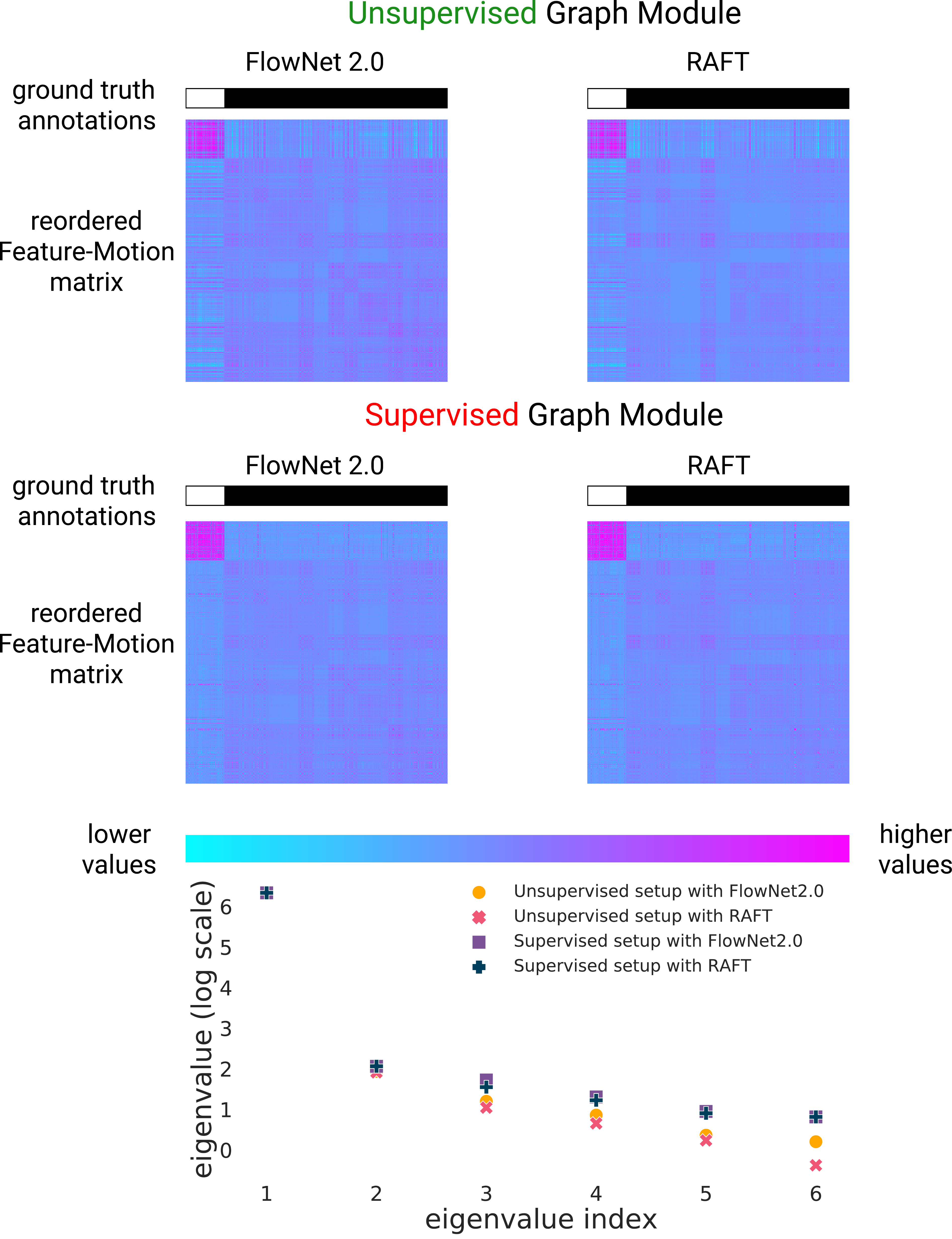}
    \centering
    \caption{Visualization of representative Feature-Motion matrices, along with their first 6 eigenvalues. We consider both the unsupervised and supervised cases and two optical flow methods. Nodes (one per matrix raw/column) are reordered such that the ones belonging to the main object of interest come first, with indices smaller than the background nodes. Above each matrix, we show the ground truth node labels (white - object; black - background). Matrices and eigenvalues are presented for the video 'car-roundabout'(DAVIS dataset), using 5 frames at a resolution of $16 \times 16$. The supervised case produces slightly stronger clusters.} 
    \label{fig:feature_motion_matrix}
\end{figure}

    The Feature-Motion matrix $\mathbf{A}$ is the key element of the proposed Graph Module. Our formulation considers segmentation as a spectral clustering problem, starting from the premise that the main object's pixels in a video sequence (in which such an object exists) form a strong natural cluster in space and time. The assumption is based on the observation that objects consist of strongly connected parts, which have a distinctive appearance distribution and display distinctive movement patterns within their space-time neighborhood. We believe that such natural properties should be captured by the space-time data structure and could also constitute the basis for initial, primal cognitive processes for object grouping. This section further tests this hypothesis by verifying that the Feature-Motion matrix contains such a main, stable space-time cluster. Such a strong cluster can then be mathematically proven to be reflected by the principal eigenvector of $\mathbf{A}$, which is the solution we propose in this work. 
    
    Therefore, we analyzed the structure of the Feature-Motion matrix. To make this analysis possible and keep computations and memory in a manageable size, we consider only the five consecutive frames in a video at a low resolution of $16 \times 16$. Note that this is necessary since for this analysis, we actually build, explicitly, the matrix $\mathbf{A}$ (as opposed to the actual algorithm for which we compute the eigenvector without having to build the matrix - a fact that would be impossible for full length, high-resolution videos).      
    In Fig.~\ref{fig:feature_motion_matrix}, we present the Feature-Motion matrices obtained in 4 configurations of our algorithm: unsupervised or supervised and with FlowNet2.0 or RAFT flow information. $\mathbf{A}_{a,b}$ models the strength of the long-range edge connecting nodes $a$ and $b$, as the Feature-Motion matrix is, in fact, the adjacency matrix of our space-time graph. Normally, nodes are indexed considering the order in which their corresponding pixels appear in the video. However, for better visualization, we have reordered the nodes according to their ground truth labels, such that the nodes belonging to the object come first, followed by the rest. This helps to highlight the natural cluster, which is formed by the nodes that truly belong to the object.  

    It can be clearly seen that in all the considered configurations, there is a very strong cluster in matrix $\mathbf{A}$, directly corresponding to the primary object, while the rest of the nodes, belonging to the background, are only weakly connected to it. Interestingly enough, note that the Feature-Motion matrices corresponding to the supervised version of our graph depict a more robust cluster than their unsupervised counterparts. This insight confirms the fact that supervised features lead to a stronger space-time clustering, which in turns leads to a better quality first eigenvector, as also confirmed in experiments (Table~\ref{tab:exp_analysis_features_unsup} and Table~\ref{tab:exp_analysis_features_sup}). We also note that, in general, there is no other strong cluster formed in the matrix besides the main, primary one. This fact becomes clear when we look at the spectrum of the matrix, namely its eigenvalues. In Fig.~\ref{fig:feature_motion_matrix} we present for each of the considered configurations, the first six eigenvalues of $\mathbf{A}$, in decreasing order. Note that there is a significant difference between the first eigenvalue and the second one for all considered setups. This difference, known as eigengap of $\mathbf{A}$, indicates that the main cluster is much stronger than the rest in the graph and is therefore stable under random perturbations. 

    It is a known result in perturbation theory~\cite{stewart1990matrix,ng2001link,leordeanu2005spectral} that for large eigengaps, we can expect the principal eigenvector to be insensitive to various errors and noises. In the ideal case, the ideal $\mathbf{A}^*$ can be considered to have strong values only in the corresponding object cluster and smaller everywhere else.  In such a case, the eigenvector will reflect the correct object cluster, having strong values for pixels belonging to the main, primary object and smaller for the rest. In practice, we expect the Feature-Motion matrix $\mathbf{A}$ to be a  perturbed version of the ideal matrix $\mathbf{A}^*=\mathbf{A}-\mathbf{E}$, where $\mathbf{E}$ is the perturbation matrix. As also shown in \cite{leordeanu2005spectral}, we can define an upper bound for any alterations of the principal eigenvector as $\epsilon\approx8\frac{\parallel\mathbf{E}\parallel_F}{\parallel\mathbf{A^*}\parallel_F}$, where $\parallel \cdot \parallel_F$ is the Frobenius norm. Thus, for relatively small perturbations $\parallel\mathbf{E}\parallel_F$, the upper bound ensures the stability of the eigenvector. 
    
     This spectral analysis of the Feature-Motion matrix confirms that the power iteration, which leads to the principal eigenvector, is a robust segmentation solution that discovers the primary object in a video as a natural cluster of points with distinctive appearance and motion patterns in space and time. Our numerical analysis also shows that the more accurate the optical flow and the more powerful the node-level object features, the stronger the cluster and the better the segmentation solution, a fact that aligns perfectly with our motivation and approach.
    
\section{Experimental analysis of the full system}
\label{sec:exp_analysis_full} 

    \subsection{Improving over several graph-network cycles}\label{sec:exp_analysis_full_system_cycles}
    \begin{figure*}[htb]
        \centering
        \includegraphics[width=1\textwidth]{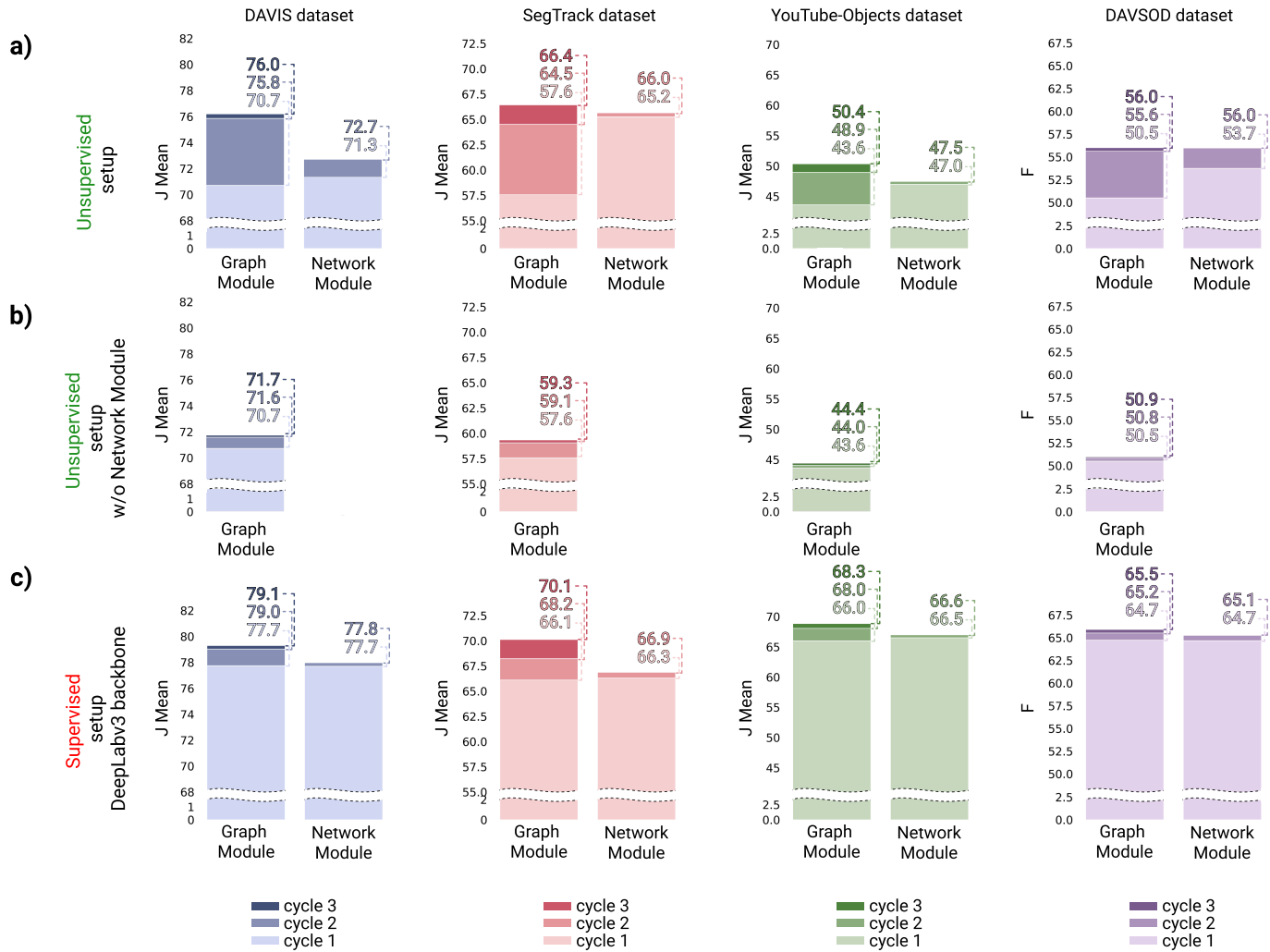}
        \caption{Performance evolution of the full Ierative Knowledge Excnahnge model. We follow both the Graph and Network Modules' evolution over several cycles. The Graph runs for an additional cycle to benefit from the best representation of the Network. \textbf{a)} the unsupervised setup \textbf{b)} the unsupervised setup without the Network Module \textbf{c)} the supervised setup over DeepLabv3 backbone. Even though the Network Module usually overcomes the Graph during the first cycle, the Graph Module exceeds the Network at convergence. The two modules' complementarity becomes evident when we consider the case without the Network (row b), with a huge performance drop compared to a). Even when starting from strong supervised features (row c), IKE still brings a significant performance boost. [Best viewed in color]}
        \label{fig:CS_GDM_results}
    \end{figure*}
    
     First, we will check the effectiveness of the iterative knowledge exchange process. We will verify that the IKE system, in which the graph functions as a teacher to the network module, and then the network provides more powerful features for the next clustering and learning cycle, is indeed effective. In Table ~\ref{tab:full_system_avgs} and Fig.~\ref{fig:CS_GDM_results} we present in detail the performance evolution over multiple datasets, considering both the unsupervised and supervised cases. From the experiments conducted so far, it should be clear that the graph is able to improve over the initial backbone network. At the same time, it is also evident that the stronger the features in $\mathbf{F}$, the better the performance. The experiments of the current section come to confirm that the graph and the network functioning in tandem, within a single knowledge-exchange system, improve over at least three cycles (outer loop in Fig.~\ref{fig:cs_gdm}) in all our experiments. In general, after 3 cycles, the system tends to converge so that neither the graph nor the net continues to improve compared to the previous cycle. It is also expected for the graph to have superior performances at convergence, as it fully benefits from the temporal dimension while the net is limited to a frame-level perspective.
    
    We now analyze the experiments in a bit more detail: in Fig.~\ref{fig:CS_GDM_results}~a), we present the performance evolution of the system in the unsupervised case, when only flow features are used at the nodes. In all experiments (supervised or not), the network module is always randomly initialized. In these experiments for the space-time graph, we consider the RAFT~\cite{teed2020raft} flow information. We observe that on all datasets the fully cyclic system achieves a remarkable performance (for comparisons to state-of-the-art see the next subsection). As also presented in Table~\ref{tab:full_system_avgs}, the unsupervised formulation displays the most significant performance boost, with up to $10.4\%$ relative percentage change between cycles.
    
    \begingroup
    \setlength{\tabcolsep}{2pt}
    \begin{table}
        \centering
        \captionsetup{justification=centering}
        \caption{\\The Relative Percentage Change Between Cycles}
        \label{tab:full_system_avgs}
        \begin{tabular}{ccccc}
        \toprule
        \multicolumn{5}{l}{\textcolor{m_green}{Unsupervised methods}} \\
        \toprule
         & \multicolumn{2}{c}{\textbf{IKE: }\textcolor{m_green}{Graph Module}} & \multicolumn{2}{c}{\textbf{IKE: }\textcolor{blue}{Network Module}} \\
        & \textcolor{m_green}{$\mathbf{x}\#1$} $\rightarrow$ \textcolor{m_green}{$\mathbf{x}\#2$} 
        & \textcolor{m_green}{$\mathbf{x}\#2$} $\rightarrow$ \textcolor{m_green}{$\mathbf{x}\#3$} 
        & \textcolor{m_green}{$\mathbf{x}\#1$} $\rightarrow$ \textcolor{blue}{$\mathbf{s}\#1$} 
        & \textcolor{blue}{$\mathbf{s}\#1$} $\rightarrow$ \textcolor{blue}{$\mathbf{s}\#2$} \\ 
        \cmidrule(lr){2-3}
        \cmidrule(lr){4-5}
         \multicolumn{1}{l}{w/ Network} & 10.4\% & 1.7\% & 7.0\% & 2.1\%  \\
        \multicolumn{1}{l}{w/o Network} & 1.3\% & 0.4\% & - & - \\
        \toprule
        \multicolumn{5}{l}{\textcolor{red}{Supervised methods}} \\
        \toprule
         & \multicolumn{2}{c}{\textbf{IKE: }\textcolor{m_green}{Graph Module}} & \multicolumn{2}{c}{\textbf{IKE: }\textcolor{blue}{Network Module}} \\
         & \textcolor{m_green}{$\mathbf{x}\#1$} $\rightarrow$ \textcolor{m_green}{$\mathbf{x}\#2$} 
        & \textcolor{m_green}{$\mathbf{x}\#2$} $\rightarrow$ \textcolor{m_green}{$\mathbf{x}\#3$} 
        & \textcolor{m_green}{$\mathbf{x}\#1$} $\rightarrow$ \textcolor{blue}{$\mathbf{s}\#1$} 
        & \textcolor{blue}{$\mathbf{s}\#1$} $\rightarrow$ \textcolor{blue}{$\mathbf{s}\#2$} \\ 
        \cmidrule(lr){2-3}
        \cmidrule(lr){4-5}
        \multicolumn{1}{l}{with DeepLabv3} & 2.2\% & 1.0\%  & 0.3\% & 0.5\%  \\
        \multicolumn{1}{l}{with FCN} & 2.8\% & 0.5\%  & 0.6\% & 0.6\%  \\
        \bottomrule
        \end{tabular}
         \parbox{\columnwidth}{\footnotesize%
\vspace{1eX}\textit{Results for the Graph ($\mathbf{x}\#c$ - graph result for cycle c) and Network Modules ($\mathbf{s}\#c$ - net result for cycle c). We consider the unsupervised case, with or without the Network Module, and the supervised case with DeepLabv3 or FCN backbone features. We show average relative percentage changes over four datasets: DAVSOD, DAVIS, YouTube-Objects, and SegTrack. The results emphasize well the larger gains obtained over the 3 cycles by the unsupervised IKE and the key role of the Network Module. The relative percentage change between two quantities $v_2$ and $v_1$ is computed as $100 * (v_2-v_1)/v_1$.}}
    \end{table}
    
    \endgroup
     
    The unsupervised formulation of our system is the most valuable since the system benefits from the clustering power of the space-time graph and the network's learning power to make learning possible without human annotations involved at any step in the process. It is clear that the two components complement each other and benefit from the supervisory signal and the knowledge (features) exchange. 
    
    During a cycle, the node-level features of the Graph Module are initialized by concatenating both its own output and the network module representation,  from the previous cycle (Alg.~\ref{alg:cs-gdm}). To validate that the performance improvement is due to the addition of high-level features from the network, we define a cyclical baseline system that is composed only of the Graph Module (depicted as w/o Network Module in Fig.~\ref{fig:CS_GDM_results}~b) and Table~\ref{tab:full_system_avgs}). In this simplified graph-only approach, the graph is initialized using only its own output from the previous cycle. In the absence of the Network Module, we still observe a small improvement between cycles, but insignificant compared to our full system's performance. This shows the complementarity between our modules and the real value of using both in the system. However, it is still interesting to observe the 'self-improving' property displayed by the graph-only module, which can exploit its previous output as an additional signal to build a more robust representation of the primary object. 

     To test the proposed approach's robustness, we also consider the supervised scenario, where the Graph Module benefits from supervised features during the first cycle. The network module, as before, is still randomly initialized and trained from scratch for each video sequence. In Fig.~\ref{fig:CS_GDM_results}~c) we present the results of our experiment considering supervised features from DeepLabv3 backbone and RAFT~\cite{teed2020raft} flow information. In Table~\ref{tab:full_system_avgs}, we present the average relative percentage changes for both DeepLabv3 configuration and a configuration using FCN features. The improvement between cycles is lower than the one observed in the unsupervised scenario, which is expected behavior as the system starts with an informed prior and reaches competitive results from the first run of the Graph Module. Note that the biggest improvement is observed on the SegTrack dataset, which also includes videos of objects not usually present in segmentation datasets, making the pretrained prior of the supervised case less relevant. Our supervised experiment validates the importance of the graph discovery module, highlighting that our system can cope with different classes of objects, in contrast to heavily supervised methods that are often highly specialized and also limited to only a few, well-known common object classes.
    
\subsection{Comparisons to baselines and state-of-the-art}
\label{sec:exp_analysis_quant_qual_comparison}

    \begin{figure*}[htb]
        \centering
        \includegraphics[width=1\textwidth]{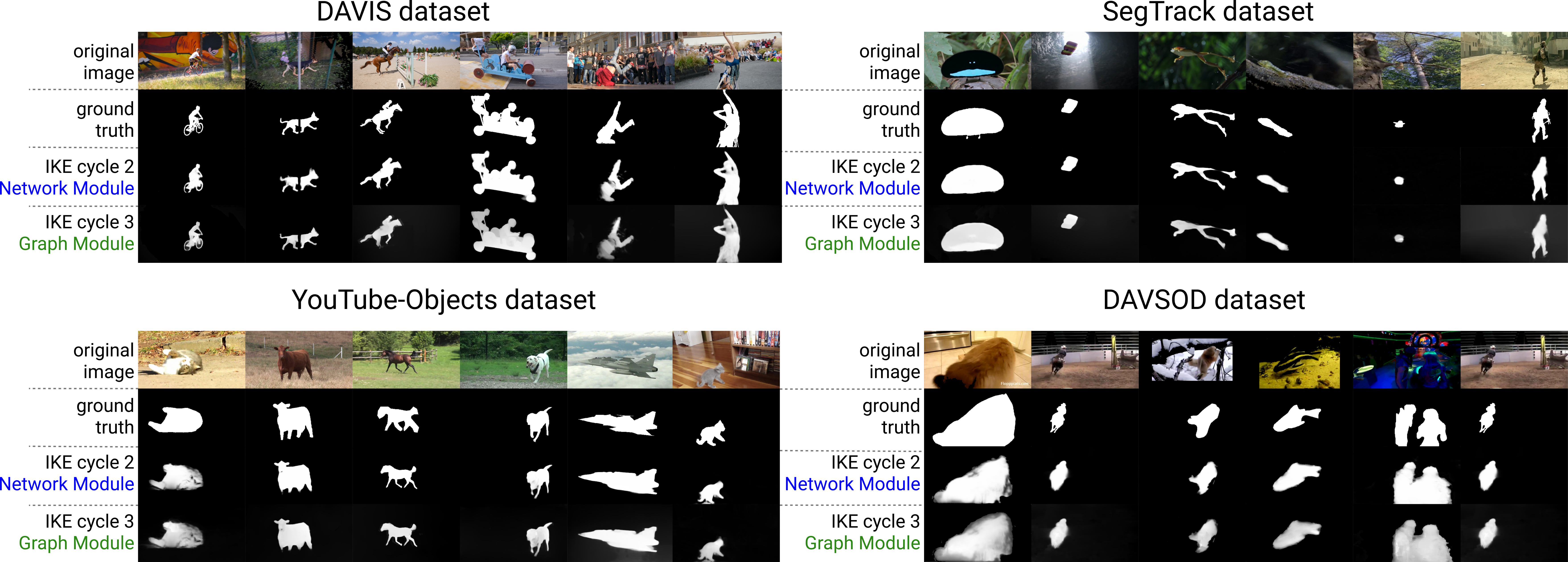}
        \caption{Qualitative results of our \textbf{ \textcolor{m_green}{unsupervised}} system, for both Network and Graph Modules for all 4 datasets considered. For YouTube-Objects and DAVSOD, the ground truth is sometimes rough, and our results tend to be, in those cases, more refined than the annotations. This emphasizes the difficulty of obtaining highly accurate human annotations.}
        \label{fig:qualitative_samples}
    \end{figure*}
    
    \begin{figure}[htb]
        \centering
        \includegraphics[width=1\columnwidth]{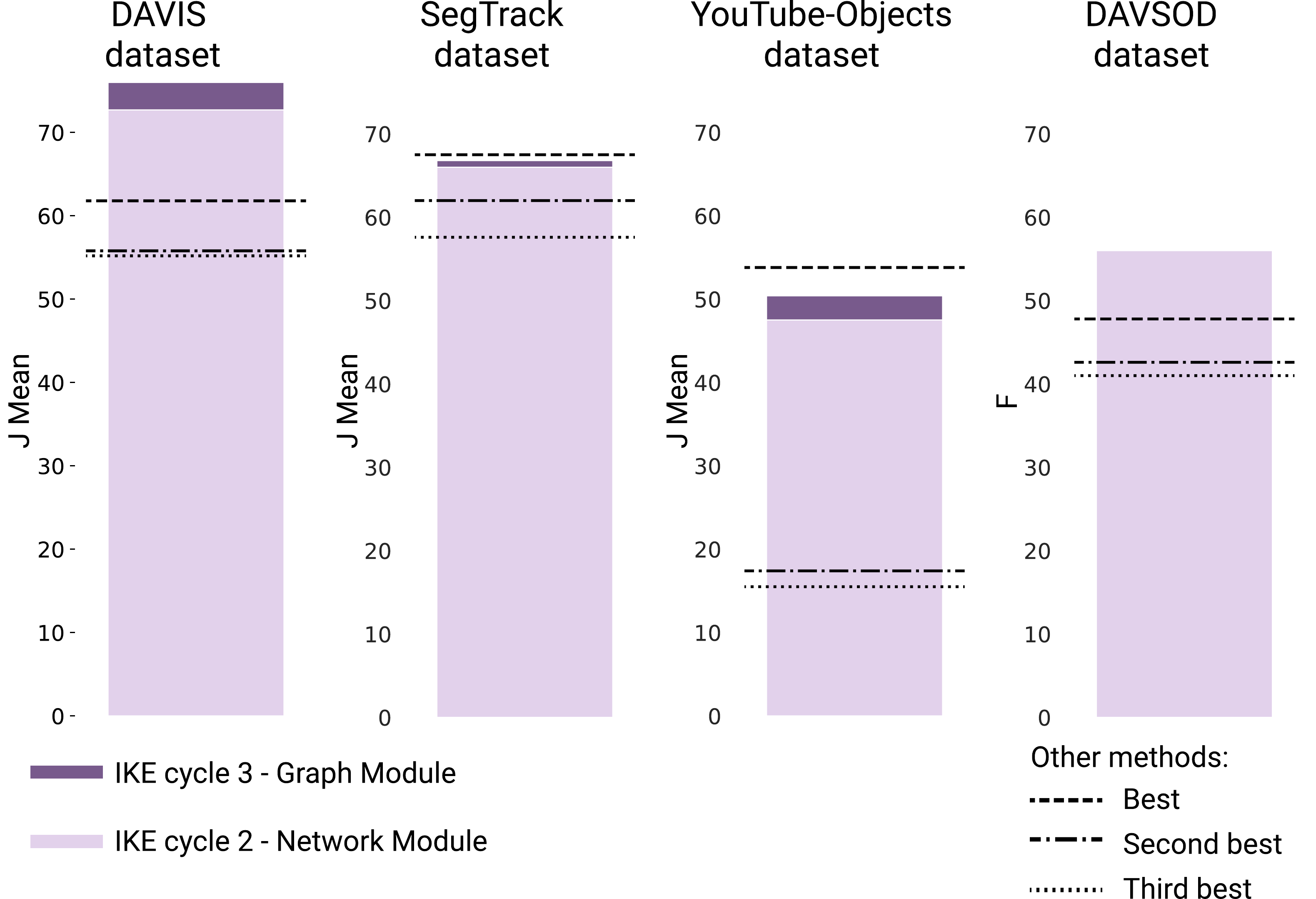}
        \caption{Final performance of Graph ($3^{rd}$ cycle) and Network ($2^{nd}$ cycle) Modules for our unsupervised IKE model (for DAVSOD, both modules have the same performance). For each dataset, we also report the results of the top 3 published methods, which also do not use any human annotations in training or pretraining. In all cases we are either ranked $1^{st}$ or $2^{nd}$.  [Best viewed in color]}
        \label{fig:quantitative_comp_unsup}
    \end{figure}
    
    \begin{figure}[htb]
        \centering
        \includegraphics[width=1\columnwidth]{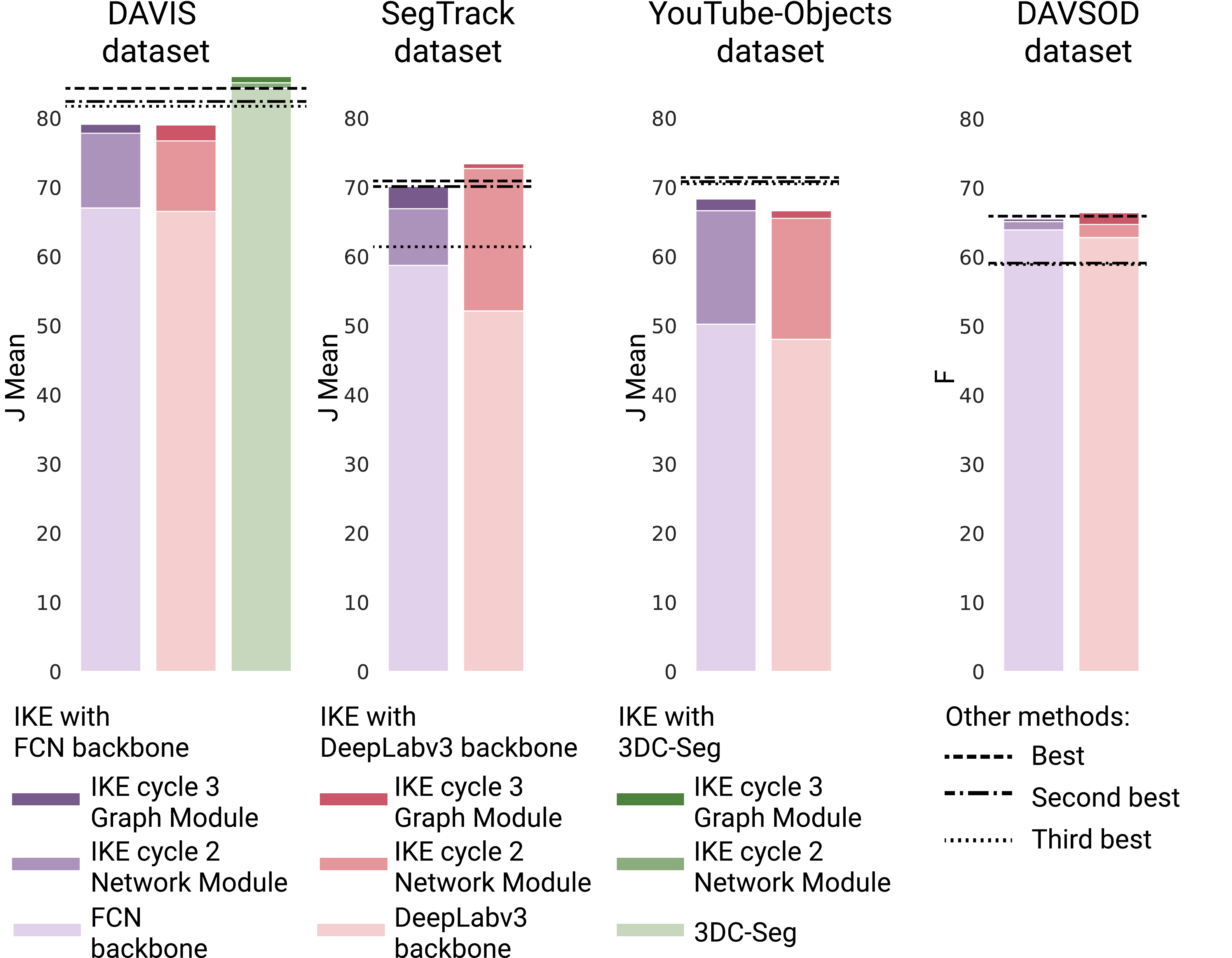}
        \caption{Final performance of Graph and Network Modules in the supervised IKE, with DeepLabv3, FCN or 3DC-Seg features. We also report results for the top 3 supervised methods. In 3 out of 4 cases, we rank $1^{st}$ [Best viewed in color]}
        \label{fig:quantitative_comp_sup}
    \end{figure}
    
    \begin{table}[t]
        \centering
        \captionsetup{justification=centering}
        \caption{\\Quantitative Comparison on DAVSOD Dataset \\ for the Video Salient Object Detection Task}
        \label{tab:davsod}
        \begin{tabular}{ll cc}
        \toprule
         \multirow{9}{*}{\rotatebox[origin=c]{90}{\parbox{1.5cm}{{\textcolor{m_green}{{\shortstack{Unsupervised\\Methods}}}}}}} &  & F $\uparrow$ &  MAE $\downarrow$ \\
        \cmidrule(lr){3-4}
        & SFLR \cite{chen2017video}     & 0.478 & 0.132 \\
        & SGSP \cite{liu2016saliency}   & 0.426 & 0.207 \\
        & STBP \cite{xi2016salient}     & 0.410  & 0.160 \\
        & SAGM \cite{wang2015saliency}  & 0.370  & 0.184\\
        & MSTM \cite{tu2016real}        & 0.344 & 0.211\\
        & MB+M \cite{zhang2015minimum}  & 0.342 & 0.228 \\
        & GFVM \cite{wang2015consistent}& 0.334 & 0.167 \\
        \cmidrule{2-4}
        &Ours & \textcolor{m_green}{\textbf{0.560}} &  \textcolor{m_green}{\textbf{0.130}}  \\
        \toprule
         \multirow{12}{*}{\rotatebox[origin=c]{90}{\parbox{1.5cm}{{\textcolor{red}{{\shortstack{Supervised\\Methods}}}}}}} & & F $\uparrow$ &  MAE $\downarrow$ \\
        \cmidrule(lr){3-4}
        &SSAV \cite{fan2019shifting}  & 0.659 & \textcolor{red}{\textbf{0.084}} \\
        & PDB \cite{song2018pyramid}        & 0.591 & 0.114  \\
        & FGRN \cite{li2018flow}            & 0.589 & 0.095  \\
        & SCNN \cite{tang2018weakly}        & 0.532 & 0.128   \\
        & DLVS \cite{wang2017video}         & 0.521 & 0.129   \\
        & MBNN \cite{li2018unsupervised}    &  0.506   & 0.109   \\ 
        & SCOM \cite{chen2018scom}          & 0.462 &  0.22\\
        & FCN backbone                      & 0.628 & 0.129 \\
        & DeepLabv3 backbone                & 0.639 & 0.122\\
        \cmidrule(lr){2-4}
        & \shortstack[l]{Ours with FCN backbone} & \textcolor{red}{\textbf{0.664}} & 0.102 \\
        & Ours with DeepLabv3 backbone & 0.655 & 0.114 \\
        \bottomrule
        \end{tabular}
        \parbox{\columnwidth}{\footnotesize%
        \vspace{1eX} \textit{Our solution ranks first in unsupervised and supervised setups, with a significant improvement over the next best solutions. (\textcolor{m_green}{\textbf{green}} - best unsupervised; \textcolor{red}{\textbf{red}} - best supervised) [Best viewed in color]}}
    \end{table}
    
    \begin{table}
        
        \centering
        \captionsetup{justification=centering}
        \caption{\\Quantitative Comparison on SegTrackv2 Dataset \\ for the Zero-Shot Video Object Segmentation Task}
        \label{tab:segtrack}
        \begin{tabular}{llcc}
        \toprule
         \multirow{8}{*}{\rotatebox[origin=c]{90}{\parbox{1.7cm}{{\textcolor{m_green}{{\shortstack{Unsupervised\\Methods}}}}}}}  & & J Mean $\uparrow$ &\\
        \cmidrule(lr){3-3}
        & NLC \cite{faktor2014video}        & \textcolor{m_green}{\textbf{67.2}} \\
        & CIS \cite{yang2019unsupervised}   & 62.0 \\
        & SAGE \cite{wang2015saliency}      & 57.6 \\
        & FST \cite{papazoglou2013fast}     & 54.3 \\
        & HPP \cite{haller2017unsupervised} & 50.1 \\
        & CUT \cite{keuper2015motion}       & 47.8 \\
         \cmidrule(lr){2-3}
        & Ours & 66.4 \\
        \toprule
        \multirow{13}{*}{\rotatebox[origin=c]{90}{\parbox{1.7cm}{{\textcolor{red}{{\shortstack{Supervised\\Methods}}}}}}} & & J Mean $\uparrow$ \\
        \cmidrule(lr){3-3}
        & EpO+ \cite{faisal2019exploiting}  & 70.9  \\
        & STP \cite{hu2018unsupervised}     & 70.1 \\
        & FSEG \cite{jain2017fusionseg}     & 61.4 \\
        & PDB \cite{song2018pyramid}        &  60.9 \\
        & IET \cite{li2018instance}         & 59.3 \\
        & KEY \cite{lee2011key}             & 57.3 \\
        & LVO \cite{tokmakov2017learning}   & 57.3 \\
        & ARP \cite{koh2017primary}         & 57.2 \\
        & FCN backbone                      & 52.1 \\
        & DeepLabv3 backbone                & 58.7\\
        \cmidrule(lr){2-3}
        & \shortstack[l]{Ours with FCN backbone} & \textcolor{red}{\textbf{73.4}} \\
        & Ours with DeepLabv3 backbone & 70.1 \\
        \bottomrule
        \end{tabular}
        \parbox{\columnwidth}{\footnotesize%
        \vspace{1eX} \textit{Our solution ranks first in the supervised setup and second in the unsupervised one. (\textcolor{m_green}{\textbf{green}} - best unsupervised; \textcolor{red}{\textbf{red}} - best supervised) [Best viewed in color]}}
    \end{table}
    
    \begin{table}[t]
        \centering
        \captionsetup{justification=centering}
        \caption{\\Quantitative Comparison on YouTube-Objects Dataset\\for the Zero-Shot Video Object Segmentation Task}
        \label{tab:ytom}
        \begin{tabular}{llc}
            \toprule
           \multirow{5}{*}{\rotatebox[origin=c]{90}{\parbox{1.7cm}{{\textcolor{m_green}{{\shortstack{Unsupervised\\Methods}}}}}}} & & J Mean $\uparrow$  \\
            \cmidrule(lr){3-3}
            & FST~\cite{papazoglou2013fast} & \textbf{\textcolor{m_green}{53.8}} \\
            & FSEG-FlowTh~\cite{jain2017fusionseg} & 17.4 \\
            & LTV~\cite{ochs2013segmentation} & 15.5 \\
            \cmidrule(lr){2-3}
            & Ours &  50.4 \\
            \toprule
            \multirow{17}{*}{\rotatebox[origin=c]{90}{\parbox{1.7cm}{{\textcolor{red}{{\shortstack{Supervised\\Methods}}}}}}} &  & J Mean $\uparrow$ \\
            \cmidrule(lr){3-3}
            & EGMN \cite{lu2020video}           & \textbf{\textcolor{red}{71.4}} \\
            & AGNN \cite{wang2019zero}          & 70.8 \\
            & COSNet \cite{lu2019see}           & 70.5 \\
            & AGS \cite{wang2019learning}       & 69.7 \\
            & FSEG \cite{jain2017fusionseg}     & 68.4 \\
            & LVO \cite{tokmakov2017learning}   & 67.5 \\
            & PDB \cite{song2018pyramid}        & 65.4 \\
            & LSMO \cite{tokmakov2019learning}  & 64.3 \\
            & COSEG \cite{tsai2016semantic}     & 58.1 \\
            & MotAdapt \cite{siam2019video}           & 58.1 \\
            & SFL \cite{cheng2017segflow}       & 57.0 \\
            & ARP \cite{koh2017primary}         & 46.2 \\
            & FCN BB                            & 48.0 \\
            & DeepLabv3 BB                      & 50.2 \\
            \cmidrule(lr){2-3}
            & Ours with FCN backbonw & 66.6 \\
            & \shortstack[l]{Ours with DeepLabv3 backbone} & 68.3 \\
            \bottomrule 
        \end{tabular}
        \parbox{\columnwidth}{\footnotesize%
\vspace{1eX} \textit{ We have competitive results in both unsupervised and supervised setups. (\textcolor{m_green}{\textbf{green}} - best unsupervised; \textcolor{red}{\textbf{red}} - best supervised) [Best viewed in color] }}
    \end{table}
    
    \begingroup
    \setlength{\tabcolsep}{5pt} 
        \begin{table}[h]
        \centering
        \captionsetup{justification=centering}
        \caption{\\Quantitative Comparison on DAVIS2016 Dataset\\for the Zero-Shot Video Object Segmentation Task}
    \label{tab:davis}
    \begin{tabular}{llccc}
        \toprule
        \multirow{6}{*}{\rotatebox[origin=c]{90}{\parbox{1.7cm}{{\textcolor{m_green}{{\shortstack{Unsupervised\\Methods}}}}}}} & & J Mean $\uparrow$ & F Mean $\uparrow$ & Avg. $\uparrow$ \\
        \cmidrule(lr){3-5}
        & ELM\cite{lao2018extending}    & 61.8 & 61.2 & 61.5 \\ 
        & FST \cite{papazoglou2013fast} & 55.8 & 51.1 & 53.5 \\
        & CUT \cite{keuper2015motion}   & 55.2 & 55.2 & 55.2 \\ 
        & NLC \cite{faktor2014video}    & 55.1 & 52.3 & 53.7 \\
        \cmidrule(lr){2-5}
        & Ours & \textbf{\textcolor{m_green}{76.0}} & \textbf{\textcolor{m_green}{69.3}} & \textbf{\textcolor{m_green}{72.7}} \\
        \toprule
        \multirow{9}{*}{\rotatebox[origin=c]{90}{\parbox{3cm}{{\textcolor{red}{{\shortstack{Supervised\\Methods}}}}}}} & & J Mean $\uparrow$ & F Mean $\uparrow$ & Avg. $\uparrow$ \\
         \cmidrule(lr){3-5}
        & 3DC-Seg \cite{mahadevan2020making} & 84.3 & \textbf{\textcolor{red}{84.7}} & 84.5 \\
        & MATNet \cite{zhou2020matnet}  & 82.4 & 80.7  & 81.6 \\
        & AnDiff \cite{yang2019anchor}  & 81.7 & 80.5 & 81.1 \\
        & COSNet \cite{lu2019see}       & 80.5 & 79.5 & 80.0 \\
        & AGS \cite{wang2019learning}   & 79.7 & 77.4 & 78.6 \\ 
        & EpO+ \cite{faisal2019exploiting}  & 80.6 & 75.5 & 78.1 \\
        & MotAdapt \cite{siam2019video}     & 77.2  & 77.4 & 77.3 \\
        & LSMO \cite{tokmakov2019learning}  & 78.2 & 75.9 & 77.1  \\
        & PDB \cite{song2018pyramid}        & 77.2 & 74.5 & 75.9 \\
        & LVO \cite{tokmakov2017learning}   & 75.9 & 72.1  & 74.0 \\
        & FCN backbone                  & 66.5 & 64.3 & 65.4 \\
        & DeepLabv3 backbone            & 67.0  & 64.8 & 65.9\\ 
        \cmidrule(lr){2-5}
        & Ours with FCN backbone & 79.0 & 74.9 & 77.0 \\
        & Ours with DeepLabv3 backbone & 79.1 & 74.9 & 77.0 \\
        & \shortstack[l]{Ours with 3DC-Seg} & \textbf{\textcolor{red}{86.0}} &  84.1 &  \textbf{\textcolor{red}{85.1}} \\
        \bottomrule
    \end{tabular}
    \parbox{\columnwidth}{\footnotesize%
\vspace{1eX} \textit{ Our system ranks first for both unsupervised and supervised cases. Note the large performance gap to the next best solution in the unsupervised case. (\textcolor{m_green}{\textbf{green}} - best unsupervised; \textcolor{red}{\textbf{red}} - best supervised) [Best viewed in color]}}
    \end{table}
    \endgroup

    In Fig.~\ref{fig:qualitative_samples}, we present qualitative visual results of our Iterative Knowledge Exchange system. We highlight the agreement between the two components, namely the Graph Module and the Network Module. An important aspect is that for both YouTube-Objects and DAVSOD, part of the ground truth annotations are rather rough. We present qualitative examples where our object masks are more refined than the actual annotations. This interesting observation highlights the need for unsupervised (or less supervised) video object segmentation solutions that can work without task-specific supervision, as it is very hard to obtain accurate human annotations.
    
    In Sec.~\ref{sec:exp_analysis_full_system_cycles}, we have presented the performance evolution of Graph and Network Modules over several cycles of our system. As illustrated in Fig.~\ref{fig:CS_GDM_results}, at convergence, the graph clearly overcomes the performance of the network, as it fully benefits from both space-time graph and deep models representation powers. Therefore, for comparison with other state-of-the-art methods, we have considered our full system's final output as the one given by the Graph Module during the third cycle. In all experiments, we consider RAFT~\cite{teed2020raft} flow information and random initialization for node labels. Also, the network module is trained from scratch for each video sequence. We make a clear distinction between supervised methods (requiring human annotations during pre-training) and unsupervised ones and test both setups for our system for all datasets considered.
    
    In Fig.~\ref{fig:quantitative_comp_unsup}, we present the final performance of both Graph Module and Network Module in the unsupervised setup (no human annotation used at any level of training or pre-training). We observe that although the graph displays superior performances, the single-image Network Module is also competitive and overcomes most of the top methods with the same level of supervision. 
    
    In Fig.~\ref{fig:quantitative_comp_sup}, we present our modules' final performance in the supervised setup.  For this experiment, we have considered both DeepLabv3 and FCN backbones. We highlight the large performance gap between our modules and the considered baselines, indicating that our algorithm cleverly exploits the space-time consistency to build a robust system.  
    
    In Tables \ref{tab:davsod}, \ref{tab:segtrack}, \ref{tab:ytom} and \ref{tab:davis}, we present a quantitative comparison of our method with other state-of-the-art solutions on DAVSOD, SegTrack, YouTube-Objects and DAVIS datasets. We report only the information relevant for the comparison.
    
     For \textbf{DAVSOD} dataset (Table~\ref{tab:davsod}), we report the results of other methods as published in \cite{fan2019shifting} and on \url{paperswithcode.com}.  In the unsupervised setup, we have state-of-the-art results with a performance gap of $0.082$ in terms of F measure concerning the next best solution. As illustrated in Fig.~\ref{fig:quantitative_comp_unsup}, at convergence, both our Graph Module and Network Module significantly overcome the other methods. In the supervised setup, our solution, using the FCN backbone, ranks first concerning the F measure, with a performance boost of $0.005$, relative to SSAV\cite{faisal2019exploiting}. 
    
    On \textbf{SegTrackv2} dataset (Table~\ref{tab:segtrack}), our solution ranks second in the unsupervised setup, with a difference of $0.8$ J Mean, concerning the first solution NLC~\cite{faktor2014video}. Although NLC overcomes our Graph Module for this dataset, our method significantly surpasses NLC on the DAVIS dataset (a boost of $20.9$ J Mean). Using the FCN backbone, we rank first in the supervised setup, with a boost of $2.5$ in terms of J Mean over the next best solution EpO+~\cite{faisal2019exploiting}. 
    
    When evaluated on \textbf{YouTube-Objects} dataset (Table~\ref{tab:ytom}), in the unsupervised scenario, our solution is only surpassed by FST~\cite{papazoglou2013fast}, which we significantly overcome on both DAVIS and SegTrack datasets. On YouTube-Objects, FST has a boost of $3.4$ J Mean over our solution, while our method overcomes FST by $12.1$ J Mean on SegTrack and $20.2$ J Mean on DAVIS. The other unsupervised methods considered for comparison are considerably weaker than ours. Our solution is placed in sixth place in the supervised setup. Still, we overcome the majority of the methods on other datasets. We also highlight that the provided annotations are not very precise. As we have presented in Fig.~\ref{fig:qualitative_samples}, there are situations when our segmentation solutions are more accurate than the provided annotations.
    
    For \textbf{DAVIS2016} dataset (Table~\ref{tab:davis}), we have considered the methods available in the official leaderboard. Our solution overcomes all the other methods by a large margin in the unsupervised setup, with a performance gap of $14.2$ J Mean over ELM~\cite{lao2018extending}, which is the next best solution. When working on top of 3DC-Seg~\cite{mahadevan2020making}, the IKE system overcomes all the considered methods.
    
    \section{Discussion and Conclusions}\label{sec:conclusions}
   We introduce a dual iterative knowledge exchange model with complementary properties. In our system, an unsupervised space-time clustering module provides supervisory signal to a deep network module, which in turn, passes back to the graph its newly learned deep features. The two complementary modules function as a single self-supervised entity and exchange information over several cycles until consensual convergence. The two parts efficiently use each other's complementary strengths to overcome their individual weaknesses. The graph clustering process transcends its initial condition and has access to powerful node-level features from the deep net, while the network receives high-quality pseudo-ground truth from the graph.  As the extensive tests show, our system can discover and segment the video sequence's primary object without requiring humanly annotated data. 

 We show that the primary object in a video can be found by computing the leading eigenvector of a special Feature-Motion matrix, which is at the core of the proposed graph module. We present an explicit mathematical model for the task of "foreground object discovery", which matches very well, as our tests also show, the ground truth given by humans. Then, besides the novel mathematical model, we also introduce an efficient algorithm for computing the huge graph's eigenvector without explicitly building its adjacency matrix, which would be computationally unfeasible. 

Our thorough experimental analysis verifies, along different dimensions, the practical and numerical properties of our approach. First, we validate the graph module's capacity to discover unsupervised, from optical flow information alone, the main object in a sequence. We test its convergence properties and motivate the proposed spectral solution by an in-depth analysis of the spectrum of the real world Feature-Motion matrices. We also demonstrate its capacity to transcend its initial backbone feature provider in the supervised case. Then we test the properties of the full cyclic graph-neural net system and show that it is effective and can self-improve significantly over three knowledge-exchange cycles. We present convincing favorable comparisons to state-of-the-art and show that our algorithm could be used in conjunction with any segmentation method as an effective refinement procedure for video segmentation. 

Overall, we prove that our approach not only makes theoretical sense but it is also effective in practice, showing that its performance matches well its underlying assumptions and claims. Our solution aligns well with the current needs of video object segmentation, as the unsupervised case becomes mandatory for developing methods that are powerful and robust on unknown data. By bringing together the complementary powers of the more classical graph clustering with modern deep learning, we reach a balance between optimization and data-driven models, an approach that could shed new light on unsupervised video segmentation research. 

\section*{Acknowledgment:} This work is funded in part by UEFISCDI, under Projects EEA-RO-2018-0496 and PN-III-P1-1.2-PCCDI-2017-0734.


\FloatBarrier
\ifCLASSOPTIONcaptionsoff
  \newpage
\fi


\bibliographystyle{IEEEtran}
\bibliography{IEEEabrv,bib}

\vskip 0pt plus -1fil

\begin{IEEEbiography}[{\includegraphics[width=1in,height=1.25in,clip,keepaspectratio]{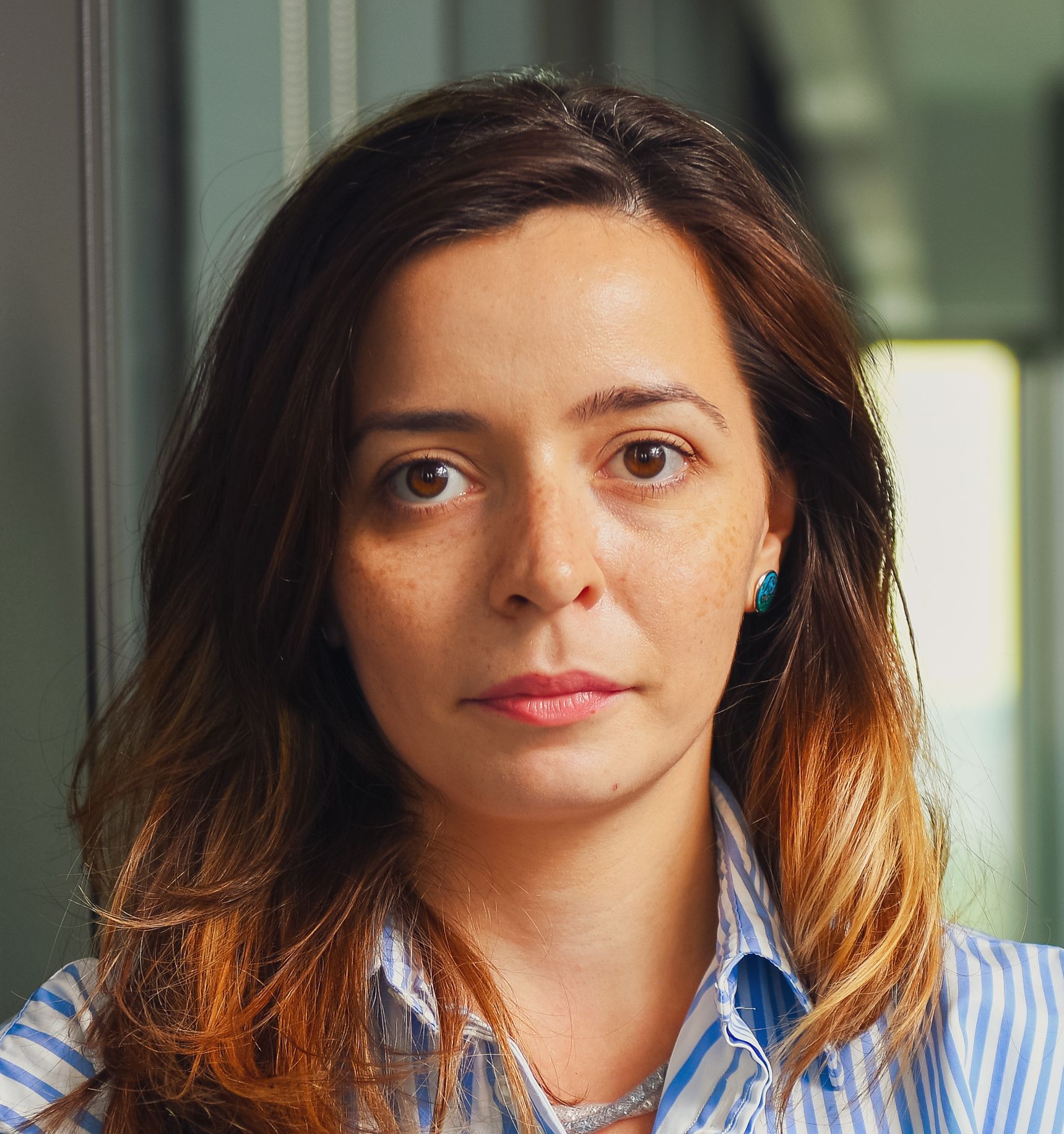}}]{Emanuela Haller}
 received her Bachelor's degree in Computer Science from University Politehnica of Bucharest and the Master's degree in Artificial Intelligence from the same institution. She has a strong background in mathematics and general computer science and  currently focusing on fundamental research. Her current work is directed towards the unsupervised video sequences analysis, focusing on zero-shot video object segmentation task. She is currently a Ph.D. student at the University Politehnica of Bucharest and part of the Theoretical Research team at Bitdefender, a global cybersecurity leader.    
\end{IEEEbiography}

\vskip -2\baselineskip plus -1fil

\begin{IEEEbiography}[{\includegraphics[width=1in,height=1.25in,clip,keepaspectratio]{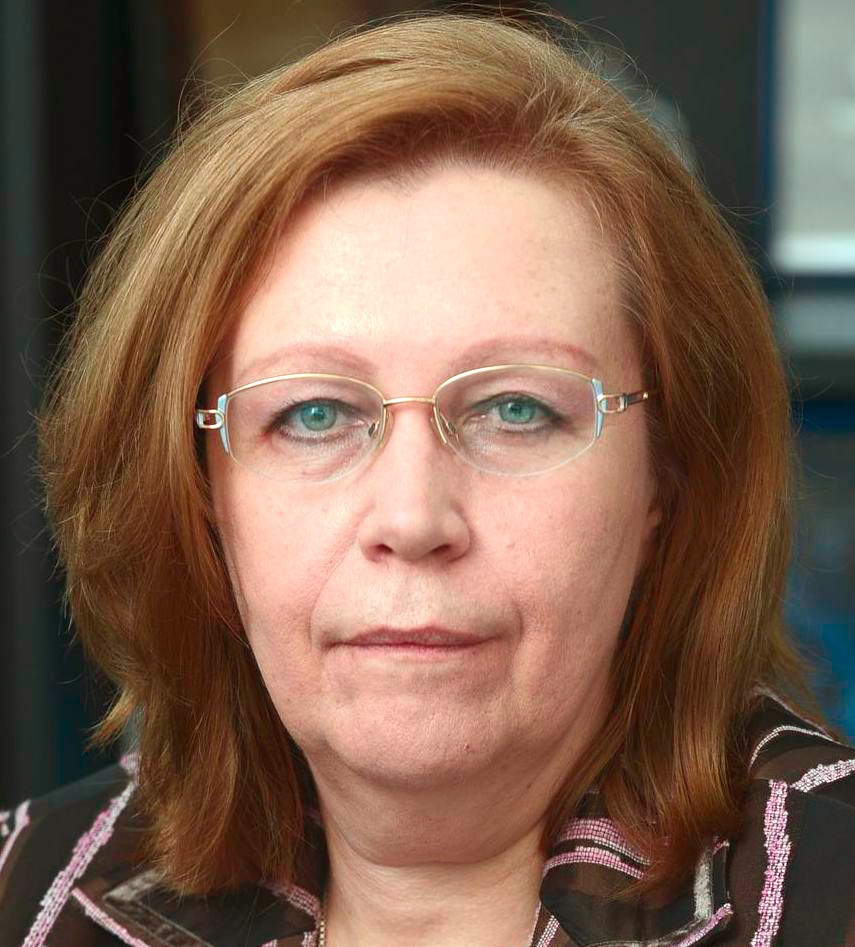}}]{Adina Magda Florea}
is Professor at the Department of Computer Science of University Politehnica of Bucharest and Head of the Artificial Intelligence and Multi-Agent Systems Laboratory (https://aimas.cs.pub.ro/). Her research interests are in multi-agent systems, machine learning, ambient intelligence, social robots and human-robot interaction. She is Senior Member of IEEE, Senior Member of ACM, and President of the Romanian Association for Artificial Intelligence.
\end{IEEEbiography}

\vskip -2\baselineskip plus -1fil

\begin{IEEEbiography}[{\includegraphics[width=1in,height=1.25in,clip,keepaspectratio]{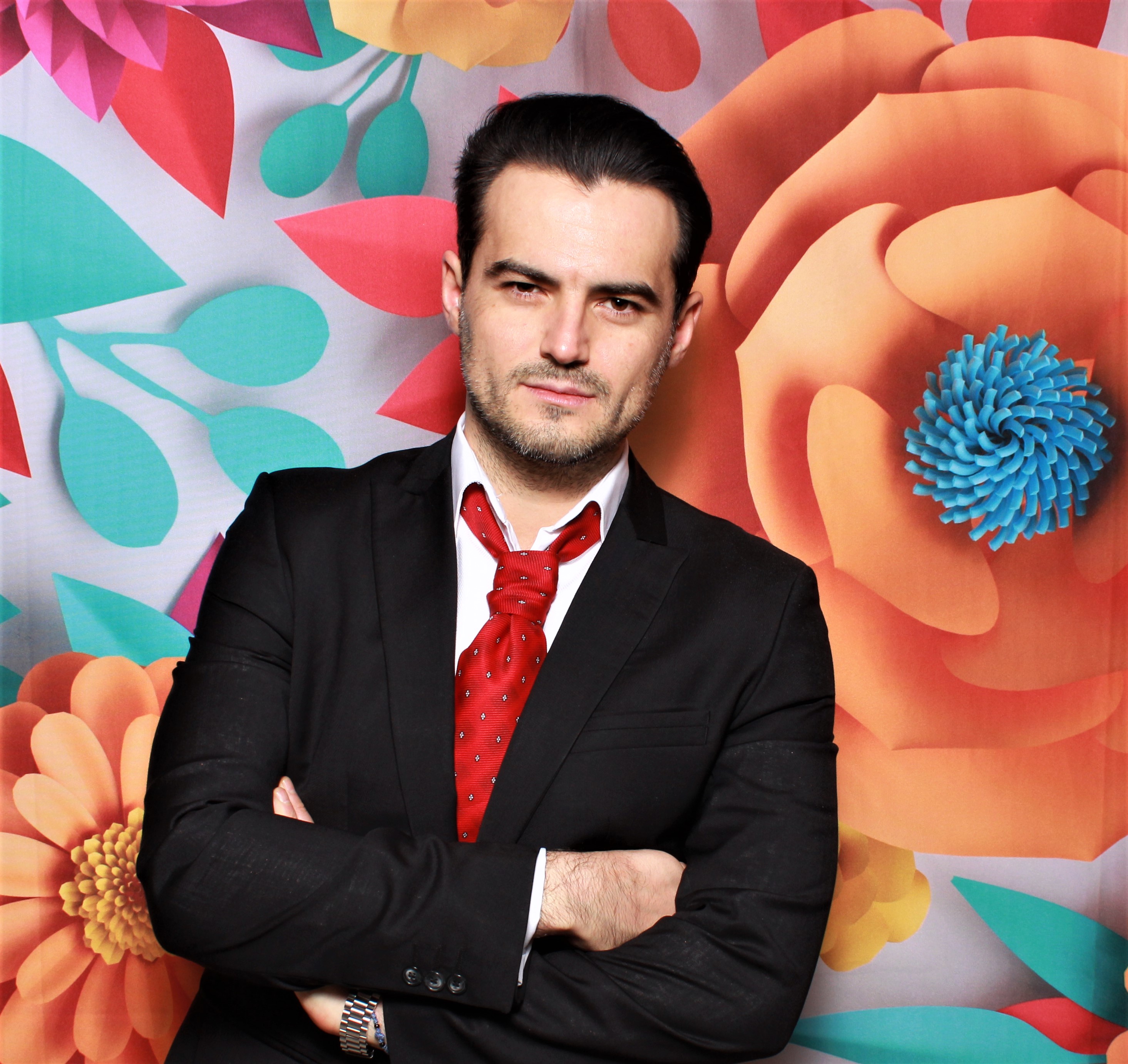}}]{Marius Leordeanu}
is Associate Professor at the University Politehnica of Bucharest (UPB) and Senior Researcher at the Institute of Mathematics of the Romanian Academy (IMAR). Marius obtained his Bachelor's in Mathematics and Computer Science at Hunter College, City University of New York (2003) and PhD in Robotics at Carnegie Mellon University (2009). At UPB he introduced the graduate courses on computer vision and robotics and at IMAR he organizes an advanced computer vision reading group with weekly meetings. His current research spans different areas in vision and learning, with focus on unsupervised learning, the space-time domain, drones and aerial scene understanding, optimization on graphs and neural nets and relating vision and language. In 2020 Marius published a book, Unsupervised Learning in Space and Time (Springer), which pushes his research towards developing a more general model for unsupervised learning in space and time. For his work on unsupervised learning for graph matching, Marius received the "Grigore Moisil Prize" (2014), the top award at the intersection of Mathematics and Computer Science, given by the Romanian Academy.
\end{IEEEbiography}




\end{document}